\documentclass{article}

\usepackage{arxiv}

\usepackage{amsmath}
\usepackage[utf8]{inputenc} 
\usepackage[T1]{fontenc}    
\usepackage{hyperref}       
\usepackage{mathptmx}
\usepackage{url}            
\usepackage{booktabs}       
\usepackage{amsfonts}       
\usepackage{nicefrac}       
\usepackage{microtype}      
\usepackage{lipsum}		
\usepackage{graphicx}
\usepackage{doi}
\usepackage[numbers]{natbib}
\usepackage{multirow}

\title{Conditional deep generative models as surrogates for spatial field solution reconstruction with quantified uncertainty in Structural Health Monitoring applications  }

\date{} 					

\author{ {\hspace{1mm}Nicholas E. Silionis \thanks{Ship-Hull Structural Health Monitoring (S-H SHM) Group,
	School of Naval Architecture and Marine Engineering,
	National Technical University of Athens,
	9 Heroon Polytechniou Av., 15780, Zografos, Athens, Greece}}
	\And
	\hspace{1mm} Theodora Liangou\footnotemark[1]
	\And
	\hspace{1mm}Konstantinos N. Anyfantis \footnotemark[1] , \thanks{Corresponding author. Tel.: +30 210 772 1325, Email: kanyf@naval.ntua.gr}}

\renewcommand{\headeright}{\emph{February 14, 2023}\hspace{92 mm}\emph{Preprint submitted to Elsevier journal}}

\renewcommand{\undertitle}{}
\renewcommand{\shorttitle}{}

\begin{document}
\maketitle

\begin{abstract}
	In recent years, increasingly complex computational models are being built to describe physical systems which has led to increased use of surrogate models to reduce computational cost. In problems related to Structural Health Monitoring (SHM), models capable of both handling high-dimensional data and quantifying uncertainty are required. In this work, our goal is to propose a conditional deep generative model as a surrogate aimed at such applications and high-dimensional stochastic structural simulations in general. To that end, a conditional variational autoencoder (CVAE) utilizing convolutional neural networks (CNNs) is employed to obtain reconstructions of spatially ordered structural response quantities for structural elements that are subjected to stochastic loading. Two numerical examples, inspired by potential SHM applications, are utilized to demonstrate the performance of the surrogate. The model is able to achieve high reconstruction accuracy compared to the reference Finite Element (FE) solutions, while at the same time successfully encoding the load uncertainty.
\end{abstract}


\section{Introduction}
\label{sec1}

Over the last few decades, significant advances in the field of computational mechanics have allowed researchers and engineers to develop high-fidelity computational models of complex physical systems. These models can efficiently predict the system's behavior under different circumstances without the need for costly and time-consuming experiments. As such, they can be used across different stages of engineering practice, from system design to decision making. This work deals with a characteristic example of such systems, namely those belonging to the field of structural analysis. These are fundamentally described using partial differential equations (PDEs) that can be solved efficiently with computational schemes such as the Finite Element Method (FEM). However, some of the most salient classes of structural analysis problems today, such as those related to uncertainty quantification (UQ), inference and optimization, require a large number of model evaluation \citep{Peherstorfer2018}, \cite{Sudret2017}. In applications where this becomes significantly high, e.g., when Monte Carlo (MC) methods are employed, or for large-scale, complex structures, the associated computational cost becomes prohibitive.
Addressing this challenge requires the development of less computationally burdensome techniques, that are simultaneously capable of reproducing the prediction accuracy of high-fidelity models, within a reasonable error margin. An appealing alternative that has emerged over recent years is the use of surrogate modeling. This refers to a class of techniques, which generally belong to the field of supervised learning, that seek to discover the governing relation between specific system inputs and outputs of interest \cite{Yang2019}. As a result, they yield a cheap to evaluate, yet accurate model that is capable of predicting the system’s response to new, previously unobserved inputs. Combined with the recent advances in machine learning (ML) and data-driven modeling in general, surrogate modeling applications have expanded rapidly and can be found across a wide spectrum of applications from different scientific disciplines, including structural analysis \cite{Papadrakakis1996,Nikolopoulos2022,Papanikolaou2022,Lindhorst2014}, material design \cite{Olivier2021} and biomedical applications \cite{Upadhyay2022}.

Most approaches to surrogate modeling follow a so-called black-box approach, which relies on making certain assumptions about the functional form of the model (e.g., linear or non-linear) and then calibrating its parameters directly using observed input-output pairs. Such methods, despite their ubiquity in various applications, are incapable of producing predictions with quantified uncertainty \cite{Yang2019}, except for Gaussian Process Regression (GPR) which relies on a probabilistic foundation \cite{Rasmussen2005}. Obtaining predictions with quantified uncertainty is particularly important in applications that include decision making tasks downstream, such as those belonging to the field of Structural Health Monitoring (SHM) \cite{Hughes2021, Vega2022}. The goal of SHM is to utilize structural response data to identify the presence of a particular damage mode of interest, determine its location and severity, and ultimately utilize this information in a prognostics framework that will enable informed decisions to be made about the operation and maintenance scheduling of the structure (Condition-based Maintenance or CBM). This is essentially a Bayesian inference problem, where one seeks to determine the (posterior) joint probability distribution of certain system parameters, i.e., damage-sensitive features, given a particular response, i.e., structural response data. This formulation, along with the inherent UQ capabilities offered by Bayesian methods (e.g., Markov Chain Monte Carlo) \cite{Nagel2016} has led to their widespread adoption in SHM applications \cite{Behmanesh2015, Behmanesh2017, Cristiani2021, Cristiani2021_2}.

A common feature of Bayesian methods for SHM is that they require a large number of forward calls to a computational model of the structure, which has led to a widespread adoption of surrogate models to alleviate the associated computational burden \cite{Cristiani2021, Cristiani2021_2, Ramancha2022, Wu2022, Kamariotis2022}. While the issue of quantifying predictive uncertainty has been acknowledged and addressed in several of these works, either using GPR (e.g., \cite{Ramancha2022, Yang2021, Parno2018}, or variational inference \cite{Vega2022}, a challenge still exists concerning the dimensionality of the data. Namely, the former suffers from the curse of dimensionality, i.e., it does not scale well to high-dimensional data \cite{Liu2018, Shi2019}, which are however required during SHM system design to facilitate optimal sensor placement (OSP). OSP is an active research area within SHM and is concerned with determining sensor designs (number, location, measured quantity) that satisfy some optimality criterion related to system performance and/or cost \cite{Yang2021, Ostachowicz2019, Colombo2022, Argyris2018, Liangou2023, Capellari2018}. The efficiency of the OSP increases with the number of potentially available sensors. In a computational setting, this translates to increasing the dimensionality of the output of the structural model, or in most cases the surrogate.

Arguably the most common approach to construct surrogate models for high-dimensional outputs utilizes the following process. First, dimensionality reduction techniques are employed to discover suitable low-dimensional representations of the high-dimensional output space. Second, a mapping between the input space and those representations is learned in a typical supervised learning framework. To generate predictions unknown inputs are mapped to the low-dimensional space and then projected to the original using an inverse transformation.  Several applications following this approach can be found both in the SHM literature \cite{Yang2021, Capellari2018} and for structural analysis in general \cite{Lataniotis2018, Guo2018, Hesthaven2018}. Principal Components Analysis (PCA) is arguably the most popular choice for dimensionality reduction, while a variety of supervised techniques can be found for the second part which offer both point estimates, such as Polynomial Chaos Expansion (PCE) \cite{Lataniotis2018} and neural networks \cite{Hesthaven2018}, as well as predictions with quantified uncertainty using GPR \cite{Yang2021, Capellari2018, Guo2018}. A particularly interesting approach can be found in Nikolopoulos et al. \cite{Nikolopoulos2022_2}, where convolutional autoencoders (CAE) are employed for dimensionality reduction. CAEs are a variant of traditional autoencoders (AE) and are comprised of the encoder and decoder, which are neural networks (with convolutional layers in CAEs) that learn to encode the data to a meaningful low-dimensional (latent) space and then decode them back to the original. CAEs are particularly well-suited to extract meaningful representations from spatial field data and since they are unsupervised, the authors of this work have combined them with a feedforward neural network.

Although these methods have proven to be effective for a wide range of problems in high dimensions, they still offer only deterministic predictions. They are therefore limited when it comes to applications concerning stochastic systems, where the stochastic parameters affecting the observed outputs include more than the system inputs. This case is common in the stochastic analysis of structures, where various sources of uncertainty coexist (e.g., material properties, geometrical imperfections, loads etc.) that influence the response quantities, yet for a particular application only some of them are of interest in the input-output system. In this work, the authors aim to treat such problems as ones of high-dimensional statistics and introduce a surrogate modeling architecture based on deep generative models to tackle them.

The term generative refers to models that include the distribution of the data itself, while in those considered deep at least some dependencies are modeled using deep neural networks \cite{Mylonas2021}. The two most common types are Variational Autoencoders (VAEs) \cite{Kingma2013} and Generative Adversarial Networks (GANs) \cite{Goodfellow2014}, of which the former will be employed in this work. VAEs have a similar structure to typical AEs described earlier, with the fundamental difference that the latent space representation is enforced by the learning process to follow a specific probability distribution. This allows for the VAE to generate realizations from the output distribution by drawing samples from the latent space distribution and passing them through the trained decoder. Furthermore, the latter can be conditioned on input parameters of interest to create a Conditional VAE (CVAE), which elegantly incorporates unsupervised learning within a supervised framework.

Traditional applications of generative models can be found in computer vision \cite{Kingma2013, Goodfellow2014, Sohn2015} and speech synthesis \cite{Prenger2018}, but in recent years they have also been used in such diverse fields as the design of new molecules \cite{Gomez2018, Sanchez2018} and astrophysics \cite{Ravanbakhsh2017}. In engineering, VAEs or CVAEs have been employed to develop condition monitoring tools for wind farms \cite{Mylonas2020, Liu2022}, for early fault detection in hydrogenerators \cite{Zemouri2023} and for gas turbine engine prognostics \cite{Zaidan2016}. When it comes to surrogate modeling, existing applications mostly do not take advantage of the probabilistic foundation of generative models and simply utilize their powerful high-dimensional representation capabilities. Such examples include using CVAEs as surrogate models for transient incompressible flows \cite{Akkari2022} and a conditional GAN to predict the welding deformation field of butt-welded plates \cite{Yi2023}. Notable exceptions include the probabilistic surrogate for fatigue life estimation of wind turbine blades in Mylonas et al. \cite{Mylonas2021} and the more general exposition contained in Yang et al. \cite{Yang2019}.

In the present work, the authors have been motivated by the need, arising particularly in SHM applications, for surrogate models that can offer predictions on high-dimensional response quantities with quantified uncertainty. To that end, a probabilistic surrogate modeling architecture is proposed, based on CVAEs, that can approximate high-dimensional conditional distributions of structural response quantities and drastically reduce computational effort by avoiding FE-based MCS. The surrogate will be applied on spatial field structural response data generated by static, linear Finite Element (FE) simulations. Uncertainty will be propagated through the structural loading and the CVAE will be conditioned on structural geometry parameters. Convolutional layers will be employed within the CVAE to exploit the spatial ordering of the response. The proposed surrogate efficiently combines powerful unsupervised learning techniques within a supervised framework to reconstruct the spatial distribution of structural response quantities, while at the same time exploiting the probabilistic underpinnings of CVAEs for uncertainty quantification. To the authors’ knowledge, this is the first time that the full capabilities of deep generative models have been utilized for surrogate modeling of complex stochastic structural systems.

The paper is organized as follows: In Section \ref{sec2}, the basic notions of dimensionality reduction and deep neural networks are presented along with the basic principles of variational inference and ultimately VAEs and CVAEs. In Section \ref{sec3}, the proposed surrogate modeling scheme is introduced in detail. In Section \ref{sec4}, two numerical examples employed to assess the performance of the surrogate are presented along with their associated results and finally in Section \ref{sec5} remarks on the findings are provided along with a discussion on possible extensions.

\section{Theoretical background}
\label{sec2}

In this section, some necessary theoretical background is provided to facilitate the reader’s understanding of the methods employed in this work. A hierarchical exposition is employed, starting from the basic principles of dimensionality reduction and neural networks, and culminating with VAEs and CVAEs. It should be noted at this point that in the following all introduced variables are nominally multi-dimensional vectors unless noted otherwise.

\subsection{Autoencoders and deep neural networks}
\label{sec21}

In many engineering applications the data of interest is often high-dimensional and therefore, it is common to seek meaningful lower-dimensional representations that facilitate visualization and interpretation. As indicated by the discussion in the previous section, these representations are often used in further tasks, such as regression within the context of surrogate modeling. The term most often used to describe this process is dimensionality reduction and the different techniques employed to achieve this nominally belong to the broader family of unsupervised learning.

\begin{figure}[htp!]

	\centering
	\includegraphics{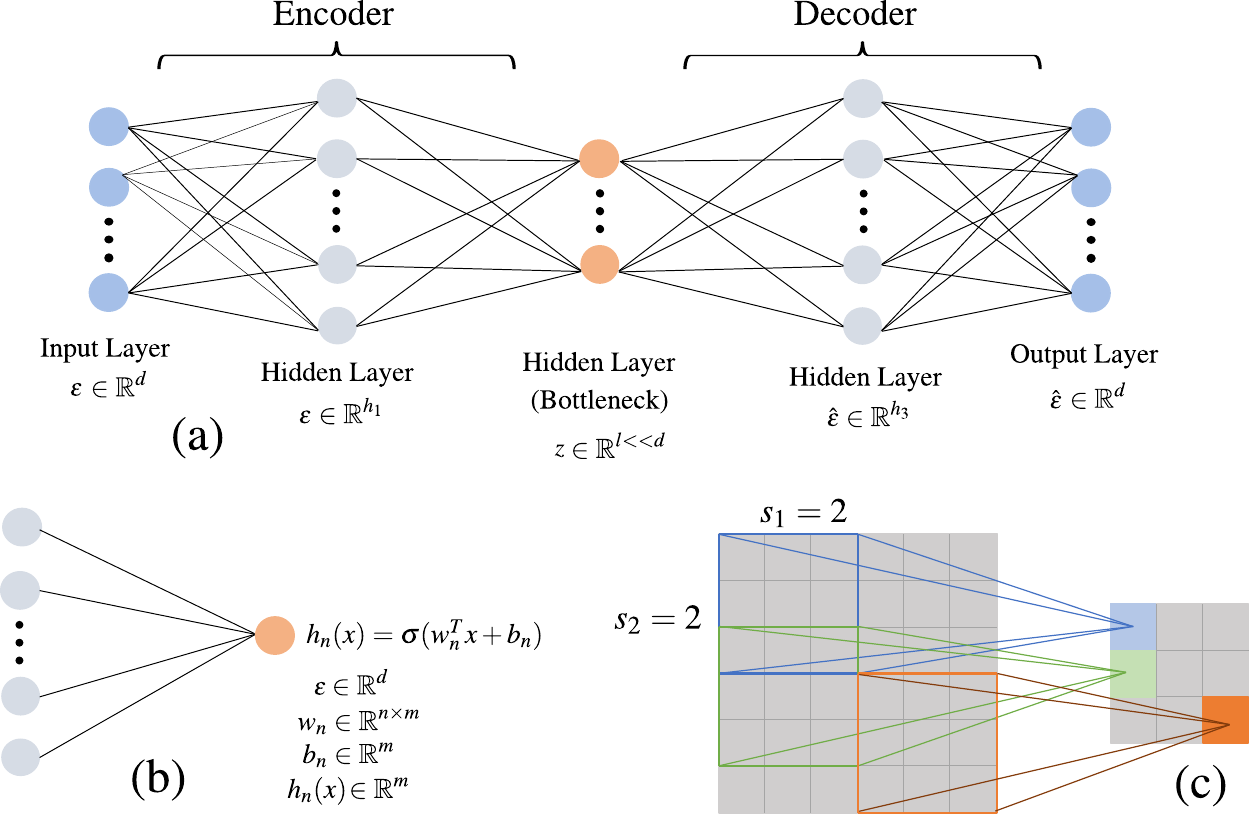}
	\caption{Schematic representation of typical (a) autoencoder with one hidden layer, (b) artificial neuron and 					(c) 2D convolution kernel}
	\label{fig: Fig 1}
	
\end{figure}

As mentioned, PCA is a classical dimensionality reduction technique that applies a linear projection to the data to transform them to a lower-dimensional space \cite{Jolliffe2011}. In many cases however, a linear transformation is incapable of providing meaningful representations and as a result, several non-linear techniques have emerged in unsupervised learning, such as kernel PCA \cite{Scholkopf1997}, diffusion maps \cite{Coifman2006} and, more pertinently for this work, autoencoders \cite{Hinton2006}. As indicated by the schematic representation of a typical autoencoder provided in Figure \ref{fig: Fig 1} (a), an AE consists of two parts. The first part, known as the encoder, can be viewed as a deterministic function $ f_{\phi}(\cdot) $, parametrized by $\phi$ which transforms the high-dimensional input data $\varepsilon \in {{\mathbb{R}}^{d}}$ to a lower-dimensional (latent) representation $z \in {{\mathbb{R}}^{l \ll d}}$. This function is typically a deep neural network (DNN) parametrized by its weights and biases. In the second part, the decoder, also a DNN, can be similarly viewed as a function $ f_{\theta}(\cdot) $, parametrized by $\theta$ that obtains a reconstruction $\hat{\varepsilon} \in {{\mathbb{R}}^{d}}$ of the initial high-dimensional data from the latent space representation.

\clearpage

It should be noted that the $\varepsilon$ notation employed for the data has been chosen for the sake of consistency throughout this work, as the data of interest that will be introduced in later sections correspond to mechanical strain, which is typically denoted with $\varepsilon$. Mathematically, the general form of an AE can be expressed as follows:

\begin{equation} \label{eq1}
f_{\phi}(\varepsilon): \mathbb{R}^{d} \rightarrow \mathbb{R}^{l}
\end{equation}
\begin{equation} \label{eq2}
g_{\theta}(\varepsilon): \mathbb{R}^{l} \rightarrow \mathbb{R}^{d}
\end{equation}
\begin{equation} \label{eq3}
\phi,\theta = \arg \max_{\phi,\theta}\mathcal{L}(g_{\theta}(f_{\phi}(\varepsilon)), \varepsilon)
\end{equation}

where $\mathcal{L}$ denotes the reconstruction loss. As indicated by this formulation, the mapping from the initial space to the latent space is learned jointly with the inverse mapping. This is a significant advantage of autoencoders compared to other non-linear dimensionality reduction techniques that require an additional training procedure to learn the inverse mapping, known as the pre-image problem \cite{Kwok2004}.

As mentioned, the parameters $\phi, \theta$ correspond to the parameters of the deep neural networks that comprise the autoencoder. Deep neural networks are powerful function approximators, which can be expressed mathematically as a composition of multi-dimensional non-linear functions. Using as an example the encoder part of the AE, and assuming that it is composed of $n$ layers, this can be written explicitly as:

\begin{equation} \label{eq4}
f_{\phi}(\varepsilon) = h_{n} \circ h_{n-1} \circ \dots h_0
\end{equation}

where $f \circ g = f(g(\cdot))$ is the composition operator and magnitude h represents the operation applied by a fully connected layer. It is illustrated for the simplified case where this layer has a single neuron in Figure \ref{fig: Fig 1} (b) and expressed mathematically as:

\begin{equation} \label{eq5}
h_n(x) = \sigma(w_n^Tx + b_n)
\end{equation}

where the vector $x$ is the input to the $n^{th}$ layer, the matrix $w_n$ contains the layer's weights and the vector $b_n$ is known as the layer's bias. The function $\sigma$ is known as the activation function and its role is to apply a non-linear transformation. Several types of activation functions exist (e.g., sigmoid, ReLU) and different ones may be applied to each layer.

Training the AE follows the same principles as training any DNN and is achieved by solving the optimization problem posed in Eq. \ref{eq3}. Generally, any differentiable function can be used as the reconstruction loss, e.g., the mean square error (MSE). The gradient of that function with respect to the parameters $\phi, \theta$, i.e., the weights and biases, is then computed using the backpropagation algorithm \cite{Rumelhart1986} through the aid of automatic differentiation \cite{Baydin2017}. Thus, gradient-based optimization methods, e.g., stochastic gradient descent (SGD), can be applied to train the network by updating its parameters until the loss is minimized.

Although traditional AEs are a powerful dimensionality reduction tool and a significant improvement compared to more traditional techniques, they too tend to suffer from the curse of dimensionality when it comes to very high-dimensional data. The reason is that the increase in the dimensionality of the input leads to an even more dramatic increase in the number of trainable network parameters. Oftentimes, data of interest in engineering applications are not only high-dimensional but also have particular characteristics, e.g., they are spatially ordered or sequential, which typical AEs are incapable of capturing efficiently.

To address these issues an appealing alternative has been proposed, namely convolutional autoencoders (CAEs) \cite{Masci2011}. CAEs share the same basic architecture with typical AEs, as described earlier, except that different types of layers are used in the DNNs in the encoder and decoder. Convolutional layers are now the dominant layer type in the former, which can additionally be combined with fully connected (linear) layers, pooling layers and normalization layers. In the latter, deconvolutional layers occupy the leading role and can be combined this time with unpooling layers, fully connected layers and normalization layers. Since the operation performed with a fully connected layer has been covered and no pooling and normalization layers have been employed in this work, the focus will be shifted to convolutional layers.

Convolutional layers, and convolutional neural networks (CNNs) \cite{Krizhevsky2012} in general, were developed as a better performing alternative to fully connected layers for image recognition problems. They are based around the convolution operator which, in the general case, takes as input a multi-dimensional array and applies to it a filter of specified size, known as the kernel, in a moving window fashion yielding an output often referred to as a feature map. Since in this work we are interested in spatially ordered data, the classical two-dimensional problem will be employed to explicitly describe the convolution operation. Let us consider a two-dimensional array, denoted for the sake of consistency as $\mathbf{E} = [\varepsilon_{ij}]$ which serves as the  input and to which a two-dimensional kernel $\mathbf{K} = [k_{ij}]$ of size $k_h \times k_v$ moving with horizontal and vertical strides $s_1$ and $s_2$ respectively is applied. This operation is shown schematically in Figure \ref{fig: Fig 1} (c). Mathematically, it can be expressed as:

\begin{equation} \label{eq6}
\varepsilon^{\prime}_{ij} = \sum\limits_{u=1}^{{{s}_{1}}}{\sum\limits_{v=1}^{{{s}_{2}}}{{{\varepsilon }_{mn}}\cdot {{k}_{uv}}+b}}
\end{equation}

where $\varepsilon^{\prime}_{ij}$ is the element of the feature map $\mathbf{E}^{\prime}$ that results from the convolution operation, $m = i \times s_2 + u$, $n = j \times s_2 + v$, $k_{uv}$ is the element of the kernel that provides the connection weights between input and feature map and $b$ is the bias term. Kernel parameters are not defined manually but are learned during the training process and typically more than one kernel is employed per layer, each producing a feature map. These are stacked, thus producing a three-dimensional tensor as the output of the convolutional layer.

Convolutional networks have several properties that make them uniquely suited to extract meaningful representations from spatially ordered data and also more appealing compared to fully connected networks. First, given that the size of the kernel is often much smaller than that of the input data, every input unit does not interact with every output unit, as opposed to fully connected layers. This leads to a decrease in the number of involved parameters, and it enables local patterns in the data to be learned efficiently. Furthermore, convolutional networks learn patterns that are equivariant to translations \cite{Goodfellow2016}, which means that once learned these patterns can be recognized anywhere within the data, as opposed to fully connected networks which would have to learn them anew. These properties make this architecture both more economical and more efficient.

In a typical CAE, multiple convolutional layers are stacked in the encoder to create a deep architecture. This allows the network to learn hierarchies of patterns, learning small low-level features in the first layer and progressively learning higher-level patterns made from the features of earlier layers. After a certain level of reduction is achieved, the encoded input is flattened to a vector and passed through one, or more, fully connected layers that ultimately produce its latent representation. Then, it enters the decoder where it is again passed through fully connected layers before being reconstructed by the deconvolutional layers. They essentially perform the reverse operation of convolution, following a similar process, although this time the kernel maps lower-dimensional inputs to higher-dimensional outputs. Ultimately, the learned deconvolution kernel parameters provide a basis for the reconstruction of the inputs to a desired output shape. Training the CAE amounts to solving the same optimization problem defined in general terms in Eq. \ref{eq1} – \ref{eq3}, again using the backpropagation algorithm.

\subsection{Conditional variational autoencoders}
\label{sec2_2}

The methods discussed until this point are still only capable of providing point estimates, i.e., statistical moments of the underlying multi-dimensional data distribution. On the other hand, generative models can learn that latent distribution and thus allow drawing samples directly. For the type of generative model employed herein, i.e., CVAEs, latent variable modeling is a key notion. According to this, instead of attempting to directly compute the probability distribution of the data of interest, denoted here as $p(\varepsilon)$, it is assumed that there exists a data-generating process from which the observations $\varepsilon$ are generated. It is also assumed that this process is governed by some latent random variables $z$. In stochastic structural systems, the latent variables can be thought of as a parametrization of the uncertainties that influence the structural response. A joint distribution over $\varepsilon$ and $z$ (with support $\mathcal{D}_z$) is assumed to exist, parametrized by some unknown hyperparameters denoted with $\theta$, for which the following relation holds according to elementary calculus of probabilities:

\begin{equation} \label{eq7}
p_{\theta}(\varepsilon, z) = p_{\theta}(\varepsilon|z)p_{\theta}(z)
\end{equation}

The marginal density of the observations, also known as the evidence, can then be obtained by integrating over the latent variables:

\begin{equation} \label{eq8}
p_{\theta}(\varepsilon) = \int_{\mathcal{D}_z} p_{\theta}(\varepsilon|z)p_{\theta}(z) \,dz \
\end{equation}

Now, the following maximum likelihood optimization goal can be set:

\begin{equation} \label{eq9}
\theta^{\prime} = \arg \max_{\theta}\left( \int_{\mathcal{D}_z} p_{\theta}(\varepsilon|z)p_{\theta}(z) \,dz \ \right)
\end{equation}

The set of hyperparameters obtained by solving this problem provides an approximation to the true data distribution. However, this integral is oftentimes intractable in closed form or requires exponential time to compute \cite{Blei2017}. Variational inference presents a clever means to overcome this problem by deriving a tractable lower bound for the quantity of Eq. \ref{eq8} using the notion of the latent variables. First, a parametric distribution over the latent variables is considered which is known as the approximate posterior and is denoted as $q_{\phi}(z|\varepsilon)$. Then, we rewrite Eq. \ref{eq8} by changing the integrand using the definition of joint probability as follows:

\begin{equation} \label{eq10}
p_{\theta}(\varepsilon) = \int_{\mathcal{D}_z} p_{\theta}(z|\varepsilon)p_{\theta}(\varepsilon) \,dz \
\end{equation}

By substituting the conditional density in the integrand with the approximate posterior and taking the logarithm of the evidence we obtain:

\begin{equation} \label{eq11}
\log \left( p_{\theta}(\varepsilon) \right) = \int_{\mathcal{D}_z} p_{\theta}(z|\varepsilon) \log \left( p_{\theta}(\varepsilon) \right) \,dz \
\end{equation}

By manipulating the term within the logarithm in the integrand using basic algebra and calculus of probabilities, which is omitted here for brevity but can be found in detail in \cite{Mylonas2021}, the following expression can be derived:

\begin{equation} \label{eq12}
\begin{split}
 \log \left( p_{\theta}(\varepsilon) \right) & = \int_{\mathcal{D}_z} q_{\phi}(z|\varepsilon)\log \left( 
 		\frac{p_{\theta}(z, \varepsilon)}{q_{\phi}(z|\varepsilon)}\right) - 
 		q_{\phi}(z|\varepsilon)\log \left( \frac{p_{\theta}(z| \varepsilon)}{q_{\phi}(z|\varepsilon)}\right) \,dz \ \\
 		& = \mathbb{E}_{q_{\phi}(z|\varepsilon)} \left[ \log \left( p_{\theta}(z,\varepsilon \right) - \log \left( q_{\phi}(z|\varepsilon) \right)\right] + D_{KL} \left( q_{\phi}(z|\varepsilon) \middle\| p_{\theta}(z|\varepsilon) \right)
 		\end{split}
\end{equation}

where $\mathbb{E}[\cdot]$ is the expectation operator and the quantity $D_{KL} \left( q_{\phi}(z|\varepsilon) \middle\| p_{\theta}(z|\varepsilon) \right)$ is the Kullback-Leibler (KL) divergence between the approximate and the true posterior distribution over the latent variables, given the observations. The KL divergence is typically used as a measure of similarity between two probability distributions and is by definition non-negative \cite{Kullback1951}. 

\newpage
Asymptotically, when it approaches zero it is implied that the approximate posterior is indeed the true posterior. By virtue of its non-negativity the following holds:

\begin{equation} \label{eq13}
\log \left( p_{\theta}(\varepsilon) \right) \geq \mathbb{E}_{q_{\phi}(z|\varepsilon)} \left[ \log \left( p_{\theta}(z,\varepsilon \right) - \log \left( q_{\phi}(z|\varepsilon) \right)\right] = \mathcal{L}(\phi, \theta ; \varepsilon)
\end{equation}

The right-hand side quantity in Eq. \ref{eq13} is known as the evidence lower bound (ELBO). Due to the non-negativity of the KL divergence, by tuning the hyperparameters $ \phi, \theta$ to maximize the ELBO, the evidence term is approximately maximized as well. Furthermore, it can be shown that this implicitly also forces the approximate posterior towards the true posterior. By further rearranging the ELBO, the following can be obtained:

\begin{equation} \label{eq14}
\begin{split}
\mathcal{L}(\phi, \theta; \varepsilon) &= \mathbb{E}_{q_{\phi}(z|\varepsilon)} \left[ \log \left( p_{\theta}(\varepsilon|z) \right) +  \log \left( p_{\theta}(z) \right) -  \log \left( q_{\phi}(z|\varepsilon) \right) \right] \\
&= \mathbb{E}_{q_{\phi}(z|\varepsilon)} \left[ \log \left( p_{\theta}(\varepsilon|z) \right) + \log \left( \frac{p(z)}{q_{\phi}(z|\varepsilon)} \right) \right] \\
&= \mathbb{E}_{q_{\phi}(z|\varepsilon)} \left[ \log \left( p_{\theta}(\varepsilon|z) \right) \right] + 
\mathbb{E}_{q_{\phi}(z|\varepsilon)} \left[ \log \left( \frac{p(z)}{q_{\phi}(z|\varepsilon)} \right) \right] \\
&=  \mathbb{E}_{q_{\phi}(z|\varepsilon)} \left[ \log \left( p_{\theta}(\varepsilon|z) \right) \right] - D_{KL} \left( q_{\phi}(z|\varepsilon) \middle\| p(z) \right)
\end{split}
\end{equation}

In this formulation, the distribution over the latent variables that appears is essentially a prior distribution whose selection is subjective. Therefore, the index $\theta$ becomes redundant and, to avoid confusion, was dropped. By examining the terms in the ELBO, it becomes clear that maximizing it consists of maximizing the expectation of the first term while minimizing the KL divergence, i.e., forcing $q_{\phi}(z|\varepsilon)$ to become the prior over the latent variables. The process by which the true data distribution is approximated using unobserved variables as a latent representation resembles neural network autoencoders, which led Kingma and Welling \cite{Kingma2013} to coin the term variational autoencoder.

In typical VAEs, the deterministic functions that transform the data to obtain both the encoder $q_{\phi}(z|\varepsilon)$ and decoder $p_{\theta}(\varepsilon|z)$ distributions are deep neural networks. Therefore, the parameters $\phi, \theta$ correspond to the weights and biases of the encoder and decoder networks respectively. Using an efficient estimate of the ELBO gradient with respect to the network parameters would allow for the SGD algorithms used to train neural networks to be employed. This can be obtained through the so-called reparameterization trick \cite{Kingma2013}, which ensures that the latent space representation follows a specific probability distribution, while at the same time the ELBO is differentiable with respect to $\phi$.

To achieve this, a careful choice of $q_{\phi}(z|\varepsilon)$ is required, such as a diagonal Gaussian with mean and variance that are outputs of the encoder neural network. This makes them automatically differentiable with respect to $\phi$, and allows for samples from this distribution to be obtained using samples from the standard normal distribution by employing the following scheme, i.e., the reparameterization trick:

\begin{equation} \label{eq15}
\begin{gathered}
e \sim \mathcal{N}(0, I_{l \times l}) \\
z \sim q_{\phi}(z|\varepsilon) = \mathcal{N}(\mu_{\phi}(\varepsilon), \sigma_{\phi}^2(\varepsilon
) \cdot I_{l \times l}) = \mu_{\phi}(\varepsilon) + e \odot \sigma_{\phi}^2 (\varepsilon)
\end{gathered}
\end{equation}

where $e$ and $z$ are $l$-dimensional vectors, $\mu_{\phi}(\varepsilon)$, $\sigma_{\phi}^2(\varepsilon
)$ are the mean and variance vectors that are the outputs of the encoder, $I_{l \times l}$ is the $l$-dimensional identity matrix and $\odot$ denotes component-wise multiplication. This transformation enables the expectations that appear in the ELBO (see Eq. \ref{eq14}) to be evaluated using samples from the standard normal distribution. At the same time, the gradient of the ELBO with respect to the network parameters can be evaluated deterministically and separately for each sample. Finally, by choosing a spherical unit Gaussian prior $p(z)$ and a diagonal Gaussian reparametrized posterior $q_{\phi}(z|\varepsilon)$ the KL divergence is analytically tractable, yielding the following statistical estimator from Kingma et al. \cite{Kingma2013}:

\begin{equation} \label{eq16}
\mathcal{L}\left( \phi ,\theta ;\varepsilon  \right)\simeq \frac{1}{2}\sum\limits_{j=1}^{J}{\left( 1+\log \left( \sigma _{\phi }^{\left( j \right)}{{\left( \varepsilon  \right)}^{2}} \right)-{{\left( \mu _{\phi }^{\left( j \right)}\left( \varepsilon  \right) \right)}^{2}}-\left( \sigma _{\phi }^{\left( j \right)}{{\left( \varepsilon  \right)}^{2}} \right) \right)}+\frac{1}{L}\sum\limits_{l=1}^{L}{\log {{p}_{\theta }}\left( \varepsilon |{{z}_{l}} \right)}
\end{equation}

\begin{figure}[htp!]

	\centering
	\includegraphics[scale=0.8]{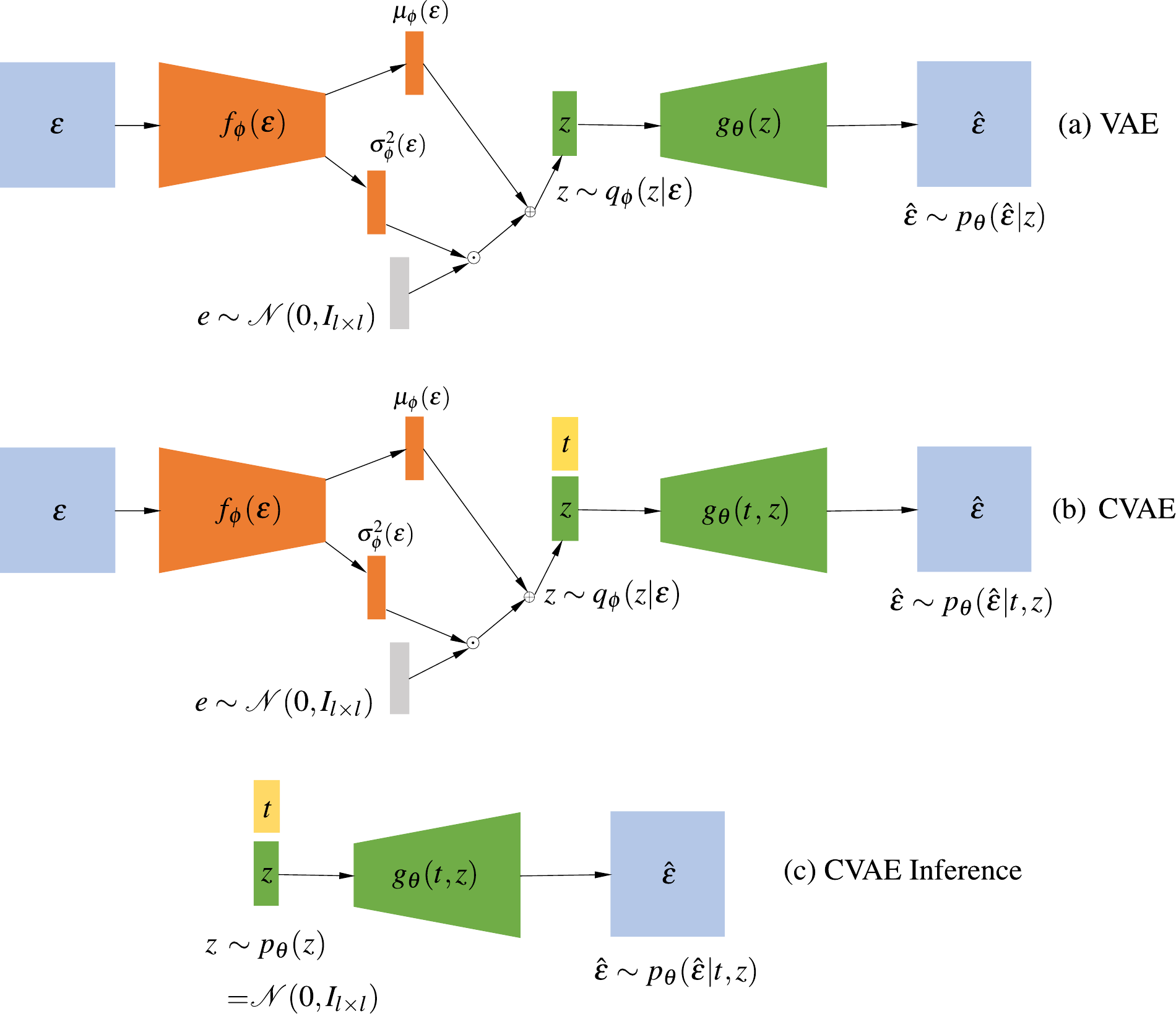}
	\caption{Computational high-level architecture of (a) VAE and (b) CVAE during training and (c) CVAE inference process}
	\label{fig: Fig 2}
	
\end{figure}

The terms of the KL divergence within the first sum are calculated element-wise over the mean and variance vectors of the $J$-dimensional Gaussian latent space, while the second sum can be interpreted as the reconstruction loss of the autoencoder, which can be calculated in principle even when a single point ($L$=1) from the latent space is sampled. It should be noted that this is not the only option in obtaining an analytical KL term. A schematic representation of a VAE computational graph is provided in Figure \ref{fig: Fig 2} (a), where the encoder and decoder neural networks are symbolized as deterministic functions using the notation of Section \ref{sec21} and the reconstructed data are denoted with $\hat{\varepsilon}$, following the common convention for neural network predictions.

When a VAE has been trained effectively, it follows from the definition of the ELBO in Eq. \ref{eq14} that the approximate posterior over the latent variables is very similar to the prior. Considering that the latter is a known distribution, generating samples from the approximated distribution of the data is straightforward. Namely, one draws a sample from the prior and then passes it through the trained decoder. Although samples drawn from this process reflect the uncertainty contained in the original data, through its encoding in the latent space distribution, the method is still unsupervised in the sense that no input-output relationship has been learned through the training process. This can be achieved using conditional VAEs, the type of architecture that is at the heart of the surrogate model proposed in this work.

A CVAE can be efficiently illustrated through its computational graph, provided in Figure \ref{fig: Fig 2} (b), where one can easily observe that it differs from the typical VAE only slightly. To explain how, let us first consider an input-output pair $\{t,\varepsilon \}$ that is typical of supervised learning, where $t$ is the $k$-dimensional vector of conditioning variables (in this work structural geometry parameters) and $\varepsilon$ denotes the structural response data. Note that the latter still contains uncertainty that has been propagated to it through other sources, not just the conditioning variables.

Conditioning is achieved by modifying the output from the latent space prior to it being fed to the decoder. This is achieved by simply stacking the conditioning variables to the reparametrized latent space representation. Then, during the training process the conditional relationship is learned by the parameters of the decoder, primarily through the reconstruction loss term in the ELBO. At the same time, the KL term enforces the latent space representation to follow a specific distribution, as dictated by the prior, which allows for the uncertainty that is contained in the data, but is not related to the conditioning variables, to be successfully encoded.

Finally, performing inference using the CVAE follows the same principle as generating data from the VAE. A schematic representation of it is provided in Figure \ref{fig: Fig 2} (c), for the case where a unit spherical Gaussian prior has been selected. To draw samples, one needs to sample from that prior, concatenate the drawn vector with that of the conditioning variables and pass it through the trained decoder to obtain a sample. Evidently, the quality of the approximation is highly dependent on the KL divergence achieved during training, which quantifies how close the approximate posterior is to the assumed prior.

\section{Conditional deep generative surrogate models for high-dimensional structural simulations}
\label{sec3}

The field of structural analysis has evolved to a point where most problems of interest are characterized by such levels of complexity, e.g., in terms of geometry, scale, loading etc., that their analytical treatment has become virtually impossible. This has led to the proliferation of computational methods, chief amongst which is the FE method, which is a numerical method used to approximate the solution of boundary and initial value problems governed by PDEs. For steady state problems, such as those considered in this work, this is achieved through an appropriate discretization of the continuous spatial variables which ultimately produces a linear system of equations that can be solved using numerical linear algebra. The obtained (discrete) solutions offer an approximation of the spatial distribution of response quantities of interest over the entire, or some sub-domain, of the structure. Since a detailed exposition of the FE method is out of the scope this work, an interested reader is referred to classic textbooks, e.g., \cite{Zienkiewicz2013, Bathe2014}.

Inherently FE solutions are high-dimensional data structures, whose dimensionality increases with the complexity of the structural problem, as does the time required to obtain them considering that each solution is associated with a matrix inversion. As such, for complex or large-scale problems the solution time associated even with a single analysis is non-trivial. This is further compounded when structural problems require probabilistic treatment in order to quantify the effects of uncertainties associated with a number of their parameters. The most versatile way to achieve this is through brute force Monte Carlo simulation, which can be broadly described as follows.

First, the basic random variables, i.e., the sources of uncertainty, of the particular problem are identified and described through their joint probability distribution. Then a large number of realizations of these variables is drawn from this distribution and propagated to the structural system. This consists of performing a deterministic FE analysis for each set of source random variable realizations drawn during this process and collecting the associated structural response quantities of interest. These are then statistically processed to quantify their probabilistic characteristics. Despite this method being straightforward, and if certain conditions are satisfied efficient, its computational requirements scale poorly to high dimensions, either in terms of the response dimensionality or the number of iterations required.

\begin{figure}[htp!]

	\centering
	\includegraphics[width=141.6 mm, height = 162.4 mm]{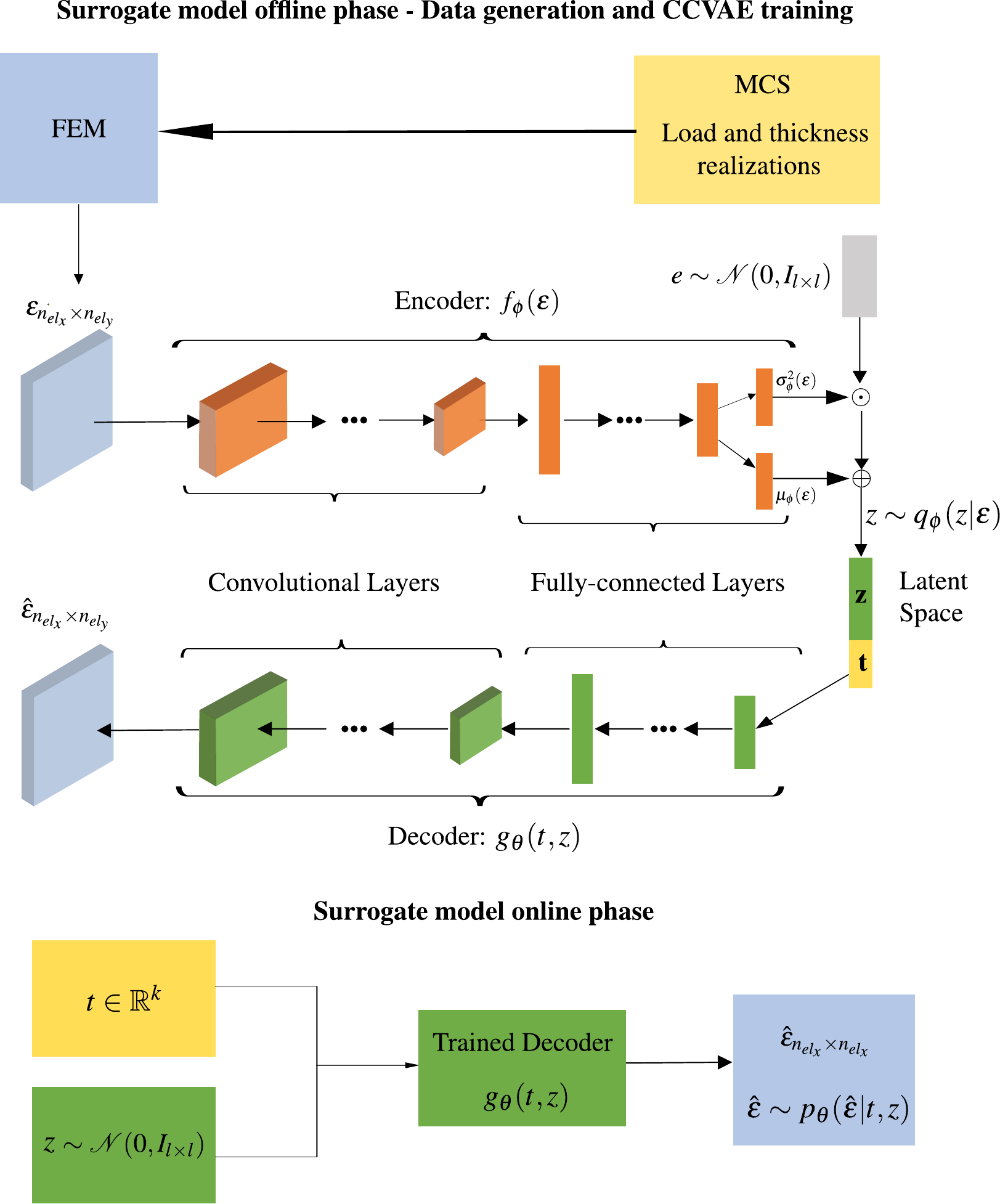}
	\caption{Schematic representation of the offline (training) and online (inference) phases of the proposed CNN-CVAE surrogate model}
	\label{fig: Fig 3}
	
\end{figure}

The goal of this work is to propose a conditional deep generative model as a surrogate that is capable of accelerating this process for a certain class of structural problems, while simultaneously not sacrificing the accuracy of the FE solutions and retaining the probabilistic nature of the analysis. The class of problems it is suitable for are stochastic static structural analyses, where spatially ordered response quantities are the desired output. The surrogate consists of a CVAE that utilizes CNNs in its encoder and decoder, which will be referred to as CNN-CVAE for the sake of brevity. A schematic representation of the proposed architecture is provided in Figure \ref{fig: Fig 3}, illustrating both the offline (training) and online (inference) stages.

This architecture offers several advantages. First, the use of CNNs leverages their powerful dimensionality reduction capabilities for spatially ordered data to efficiently reconstruct the high-dimensional FE response fields. Second, since it rests on a probabilistic foundation it is capable of quantifying predictive uncertainty. Third, its use of latent variable modeling and conditional structure allow it to efficiently approximate the desired input-output relationship and at the same time encode uncertainties contained in the data whose source is not the input parameters. This flexibility cannot be offered by traditional black-box type surrogates and is uniquely suited to surrogate models for uncertainty quantification.

The two problems considered in this work to demonstrate the surrogate model’s performance, which will be described in detail in Section 4, feature thin plate-like structural elements subjected to stochastic loading in the form of lateral pressure. The conditioning variable in both cases, i.e., the surrogate input parameter, is the plate thickness. This choice was made based on the intended use of the surrogates in SHM applications, where monitoring corrosion-induced thickness loss through strain sensing is investigated (see \cite{Liangou2023, Silionis2023}) and plate thickness acts as the damage-sensitive parameter. The two problems featured contain both a case where the conditioning variable is a scalar quantity and one where it is a vector.

\clearpage

To obtain data for training and evaluating the performance of the surrogate, a relatively small, yet representative, number of load and thickness realizations will be generated and then propagated to the FE model within an MCS. The number of realizations will be selected so as to effectively span the space both of the conditioning variables as well as the other sources of uncertainty, i.e., the plate thickness and pressure loads respectively. At the same time, this number should be kept low enough so that the offline cost does not become excessive. As shown in Figure \ref{fig: Fig 3}, the result of this process is a dataset of strain response fields $\varepsilon \in \mathbb{R}^{n_{el_x} \times n_{el_y}}$. The dimensions of these arrays, denoted as $n_{el_x}$, $n_{el_y}$ are controlled by the FE discretization of the domain and are not necessarily equal. Ultimately this dataset will be split into one for training the surrogate, and one for testing its performance to previously unseen data.

In terms of the architecture itself, as illustrated in Figure \ref{fig: Fig 3}, the encoder consists of a series of convolutional layers followed by fully connected ones that gradually reduce the dimensionality of the input. The conditioning variables are introduced to the reparametrized latent representation and then fed to the decoder, itself consisting of a series of fully connected layers that now gradually increase the dimensionality of their inputs which are then reshaped appropriately and passed through deconvolutional layers to obtain the reconstruction. As mentioned in Section \ref{sec2}, a unit spherical Gaussian has been chosen as the prior over the latent space and the MSE has been selected as the metric for the reconstruction loss (second term in Eq. \ref{eq16}). The different hyperparameters (e.g., number of layers, batch size etc.) of the CNN-CVAE were tuned using heuristics and will be described in Section \ref{sec4} for each case study separately. To generate data from the trained surrogate, the following process is employed. The vector containing the desired thickness values is concatenated with a sample drawn from the prior and the resultant vector is passed through the decoder to obtain a conditional sample from the predictive strain distribution. A schematic representation of this process, i.e., the online phase of the surrogate, is also provided in Figure \ref{fig: Fig 3}.

To assess the performance of the trained surrogate on the test set one must account for the fact that it yields samples from the approximated underlying (predictive) data distribution. Therefore, it is meaningless to compute a 1-to-1 error between exact (FE-based) and predicted response fields for a given value of the conditioning variable. The reason this is the case is that while the FE data correspond to a specific load realization, the same is not the case for the samples drawn from the surrogate. Therefore, it was decided that the error metrics employed will be based on statistics, i.e., the mean and standard deviation, calculated over the instances contained in the test. These are defined as follows:

\begin{equation} \label{eq17}
\mu_{\text{test}}^{\text{FE}} = \frac{1}{N_{\text{test}}} \sum_{i=1}^{N_{\text{test}}} \varepsilon_{\text{FE}}^{(i)}
\end{equation}

\begin{equation} \label{eq18}
\sigma_{\text{test}}^{\text{FE}} = \left[ \frac{1}{N_{\text{test}}} \sum_{i=1}^{N_{\text{test}}} \left( \varepsilon_{\text{FE}}^{(i)} - \mu_{\text{test}}^{\text{FE}} \right)^2 \right]^{1/2} 
\end{equation}

where $\varepsilon_{\text{FE}}^{(i)}$ refers to the response field corresponding to the $i^{\text{th}}$ instance in the test set and the operations are performed in an element-wise manner, thus yielding mean and standard deviation fields. Similarly, we define the same terms for the predictions obtained from the CNN-CVAE:

\begin{equation} \label{eq19}
\mu _{\text{test}}^{\text{pred}}=\frac{1}{{{N}_{\text{test}}}}\sum\limits_{i=1}^{{{N}_{\text{test}}}}{\varepsilon _{\text{pred}}^{\left( i \right)}}=\frac{1}{{{N}_{\text{test}}}}\sum\limits_{i=1}^{{{N}_{\text{test}}}}{{{g}_{\theta }}}\left( {{t}_{i}},{{z}_{i}} \right)
\end{equation}

\begin{equation} \label{eq20}
\sigma _{\text{test}}^{\text{pred}}={{\left[ \frac{1}{{{N}_{\text{test}}}}\sum\limits_{i=1}^{{{N}_{\text{test}}}}{{{\left( \varepsilon _{\text{pred}}^{\left( i \right)}-\mu _{\text{test}}^{\text{pred}} \right)}^{2}}} \right]}^{1/2}}={{\left[ \frac{1}{{{N}_{\text{test}}}}\sum\limits_{i=1}^{{{N}_{\text{test}}}}{{{\left( {{g}_{\theta }}\left( {{t}_{i}},{{z}_{i}} \right)-\mu _{\text{test}}^{\text{pred}} \right)}^{2}}} \right]}^{1/2}}
\end{equation}

where $z_i \sim p(z)$ is drawn from the prior distribution over the latent space and operations also take place in an element-wise fashion. Ultimately, we define normalized error metrics for the mean and standard deviation between the FE-based data and the CNN-CVAE predictions as follows:

\begin{equation} \label{eq21}
{{\hat{e}}_{\mu }}=\frac{{{\left\| \mu _{\text{test}}^{\text{FE}}-\mu _{\text{test}}^{\text{pred}} \right\|}_{2}}}{{{\left\| \mu _{\text{test}}^{\text{FE}} \right\|}_{2}}}
\end{equation}

\begin{equation} \label{eq22}
{{\hat{e}}_{\sigma }}=\frac{{{\left\| \sigma _{\text{test}}^{\text{FE}}-\sigma _{\text{test}}^{\text{pred}} \right\|}_{2}}}{{{\left\| \sigma _{\text{test}}^{\text{FE}} \right\|}_{2}}}
\end{equation}

where $|| \cdot ||_{2}$ refers to the $L_2$ norm. It must be acknowledged that the terms in Eq. \ref{eq21} - \ref{eq22} depend on the sampling from the prior over the latent space which influences the terms of Eq. \ref{eq19} - \ref{eq20}. Therefore, to assess the stability of the metrics and therefore the performance of the trained surrogate, an MCS over the latent variables can be performed which will enable the calculation of statistics over the error metrics themselves. 

\clearpage

\section{Numerical examples}
\label{sec4}

The proposed surrogate is applied to two case studies that have a particular SHM interest. In the first, a clamped plate under stochastic lateral pressure is considered as an illustrative example of the method’s applicability on a less complex geometry with a significant degree of uncertainty. In the second case study, the method is illustrated on a large-scale structure, a ship hull, where the surrogate is employed in lieu of a global model to reconstruct the strain response of a particular region of interest.

\subsection{Clamped plate subjected to stochastic lateral pressure}
\label{sec41}

\begin{figure}[htp!]

	\centering
	\includegraphics{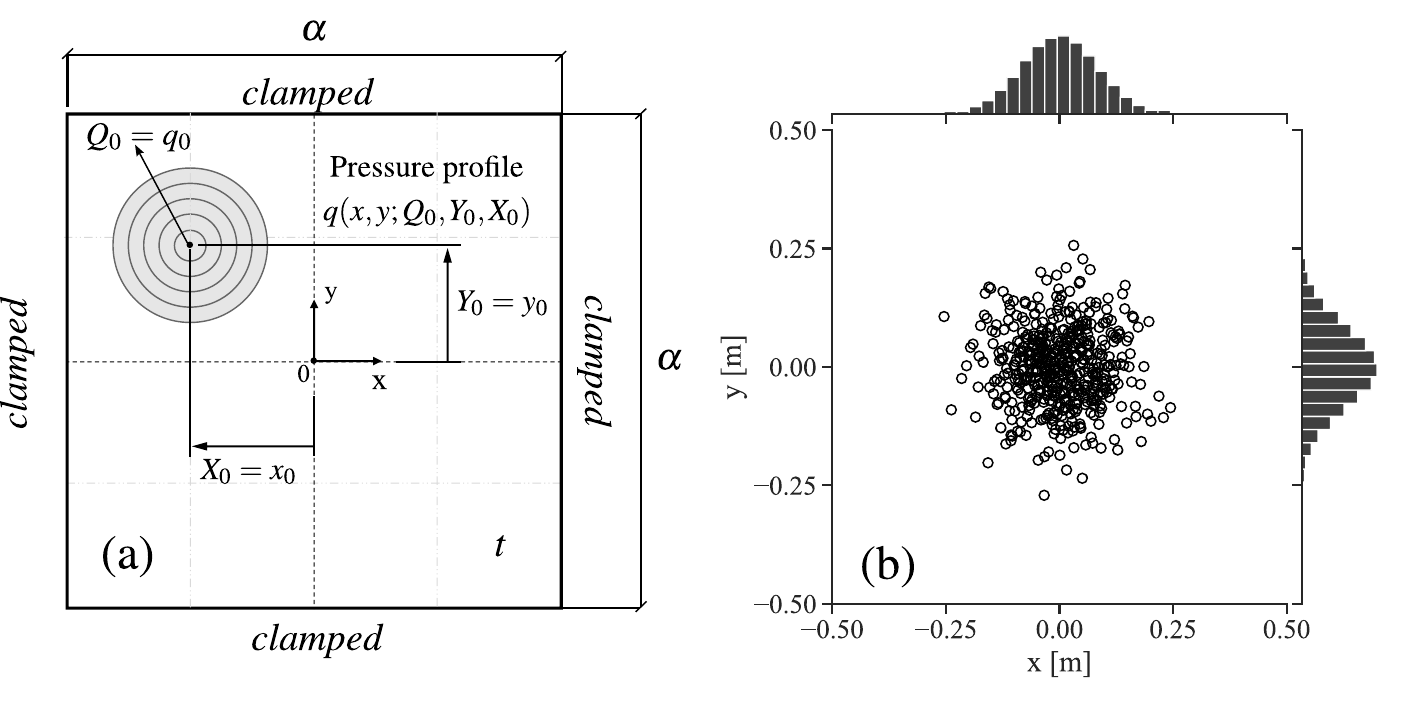}
	\caption{Schematic representation of the clamped under stochastic pressure loading (a) and indicative realizations of the peak coordinates $(X_0, Y_0)$}
	\label{fig: Fig 4}
	
\end{figure}

The first problem on which the proposed architecture was applied was a 1000×1000 mm thin steel plate, which is showcased in Figure \ref{fig: Fig 4}. Plate-like structural elements are not only the fundamental components of ship structures, which are the authors’ primary field of interest, but are ubiquitously found across engineering applications. The plate considered in this work was clamped along its edges, which translates to all degrees of freedom (DOF) along the boundary being constrained. The primary source of uncertainty was the considered structural load, a lateral pressure profile of stochastic amplitude and peak location. A schematic representation of the load profile is provided in Figure \ref{fig: Fig 4} (a). Mathematically, it can be described as follows:

\begin{equation} \label{eq23}
q(x,y;{{Q}_{0}},{{X}_{0}},{{Y}_{0}})={{Q}_{0}}\exp \left( -\frac{1}{2}{{\left( \frac{x-{{X}_{0}}}{s} \right)}^{2}} \right)\exp \left( -\frac{1}{2}{{\left( \frac{y-{{Y}_{0}}}{s} \right)}^{2}} \right)
\end{equation}

where $s$ is a deterministic shape parameter, $x$, $y$ are deterministic variables representing coordinates relative to the plate’s center, as depicted in Figure \ref{fig: Fig 4} (a), while $X_0$, $Y_0$, $Q_0$ are independent continuous basic random variables that control the load-related uncertainty. More specifically, $Q_0 \sim \mathcal{N}(q_0; \mu_{Q_0}, \sigma_{Q_0}^2)$ describes the peak amplitude of the pressure profile, while $X_0 \sim \mathcal{N}(x_0; \mu_{X_0}, \sigma_{X_0}^2)$, $Y_0 \sim \mathcal{N}(y_0; \mu_{Y_0}, \sigma_{Y_0}^2)$ describe its location. Indicative realizations of such locations are depicted in Figure \ref{fig: Fig 4} (b). The parameters of the peak amplitude distribution were selected so that the plate remains within the elastic range, and therefore the linearity of the analysis is not violated, while those of the peak coordinates $(X_0, Y_0)$ were selected so as establish that the profile remains within the plate domain with very high probability.

In terms of the FE model, the plate was discretized using 50 elements per side. Linear shell elements were selected, which contain 6 DOFs per node and are suitable for the analysis of thin plates. More information on the modeling parameters can be found in Liangou et al. \cite{Liangou2023} and will not be repeated here in the interest of conciseness. According to the process described in Section \ref{sec3}, to generate data to train and evaluate the surrogate model a forward MCS is required. For this case, $N_{\text{MC}}$ = 3000 realizations were drawn from the distributions of the independent source random variables $(X_0, Y_0, Q_0)$ using Latin Hypercube Sampling (LHS). The same number of samples was generated for the conditioning variable, i.e., the plate thickness, for two different ranges of possible values following the uniform distribution, namely $t \sim \mathcal{U}[7, 10] \ \text{mm}$ and $t \sim \mathcal{U}[7, 10] \ \text{mm}$. The uniform distribution was selected to generate realizations that effectively span the parameter space with equal probability. The two thickness ranges were chosen considering the surrogate’s intended use for thickness loss monitoring, as they represent potential thickness loss of up to 30\% and 20\% respectively. As will be shown later, they also serve to assess the surrogate’s effectiveness under different levels of uncertainty.

As a result of this process 3000 response fields $\left( \varepsilon \in \mathbb{R}^{50 \times 50} \right)$ were generated for three different strain components, namely $\varepsilon_{\text{xx}}$, $\varepsilon_{\text{yy}}$, $\varepsilon_{\text{xy}}$, which correspond to strain along the $x$ and $y$ axis, and in-plane shear strain respectively. The axis notation is the same as in Figure \ref{fig: Fig 4} (a). The analysis was carried out on a workstation with 32 GB of RAM and an Intel\textsuperscript{®} Core\textsuperscript{TM} i7-10750H CPU and took approximately 2 hours to complete. To generate the training and test sets, a 65\%-35\% split was employed, and min-max scaling was applied to normalize the training data.

The same CNN-CVAE architecture, following the principles described in Section \ref{sec3} and illustrated in Figure \ref{fig: Fig 3}, was ultimately used for all three strain components. Both this decision as well as the choice of the final hyperparameters of the surrogate, was established based on heuristics. More specifically, different combinations of hyperparameters, such as the number of layers, both convolutional and fully connected, the stride and size of the kernel in the former and number of weights in the latter, were implemented and assessed based on the training and test set performance of the model. Furthermore, different batch sizes and types of activation functions were also assessed, as well as other layer types, such as dropout and batch normalization. The final layout of the encoder and decoder of the CNN-CVAE is provided in Table \ref{tab:Tab1}.

\begin{table}[htp!]

  \centering
  \caption{Layout of CNN-CVAE architecture for the clamped plate problem}
    \begin{tabular}{cp{11.45em}p{16.05em}p{5.65em}}
    \toprule
          & Layer type & Layer Parameters & Activation \\
    \midrule
    \multicolumn{1}{c}{\multirow{8}[2]{*}{Encoder}} & 2D Convolutional & 64 (5×5) kernels with (1×1) stride & Leaky ReLU \\
          & 2D Convolutional & 128 (5×5) kernels with (2×2) stride & Leaky ReLU \\
          & 2D Convolutional & 128 (5×5) kernels with (2×2) stride & Leaky ReLU \\
          & 2D Convolutional & 64 (5×5) kernels with (2×2) stride & Leaky ReLU \\
          & Flattening layer & -     & - \\
          & Fully Connected & Weights: (1600 × 512) & Leaky ReLU \\
          & Fully Connected & Weights: (512 × 128) & Leaky ReLU \\
          & Fully Connected & Weights: (128 × 32) & Leaky ReLU \\
    \midrule
    \multicolumn{1}{c}{\multirow{8}[2]{*}{Decoder}} & Fully Connected & Weights: (33 × 128) & Leaky ReLU \\
          & Fully Connected & Weights: (128 × 512) & Leaky ReLU \\
          & Fully Connected & Weights: (512 × 1600) & Leaky ReLU \\
          & Reshaping layer & -     & - \\
          & 2D Deconvolutional & 128 (5×5) kernels with (2×2) stride & Leaky ReLU \\
          & 2D Deconvolutional & 128 (5×5) kernels with (2×2) stride & Leaky ReLU \\
          & 2D Deconvolutional & 64 (5×5) kernels with (2×2) stride & Leaky ReLU \\
          & 2D Deconvolutional & 1 (5×5) kernel with (1×1) stride & Sigmoid \\
    \bottomrule
    \end{tabular}
    
     \label{tab:Tab1}
\end{table}

As indicated by Table \ref{tab:Tab1}, the initial $50 \times 50$ input is gradually reduced to a 32-dimensional latent representation, which through the reparameterization trick is forced to follow a 32-dimensional standard normal distribution. Then through the decoder this is mapped back to the original shape, where it is worth noting that a sigmoid activation was used at the output of the last layer that provides the reconstructed strain response field. This was done to ensure that the outputs are within the $[0,1]$ range and therefore are consistent with the inputs obtained using min-max scaling.

Overall, the CNN-CVAE was trained for 100 epochs with a batch size of 8 using the Adam optimizer \cite{Kingma2014} with a learning rate equal to $10^{-4}$. The loss function of Eq. \ref{eq16} was employed with the MSE chosen as the reconstruction loss. Note that while the initial maximum likelihood goal was maximizing the ELBO, the estimator of Eq. \ref{eq16} corresponds to an equivalent minimization problem. Overall training time with GPU acceleration using an NVIDIA Quadro P620 amounted to approximately 14 minutes for each strain component. The metrics defined in Eq. \ref{eq21} and \ref{eq22} were used to assess the surrogate’s performance. Namely, they were calculated over the test instances within an MCS with $N_{\text{MC}} = 1000$ samples from the latent space distribution. The collected results are presented using histograms in Figure 5 and 6 for $t \sim \mathcal{U}[7, 10] \ \text{mm}$ and $t \sim \mathcal{U}[7, 10] \ \text{mm}$ respectively.

\clearpage

\begin{figure}[htp!]

	\centering
	\includegraphics{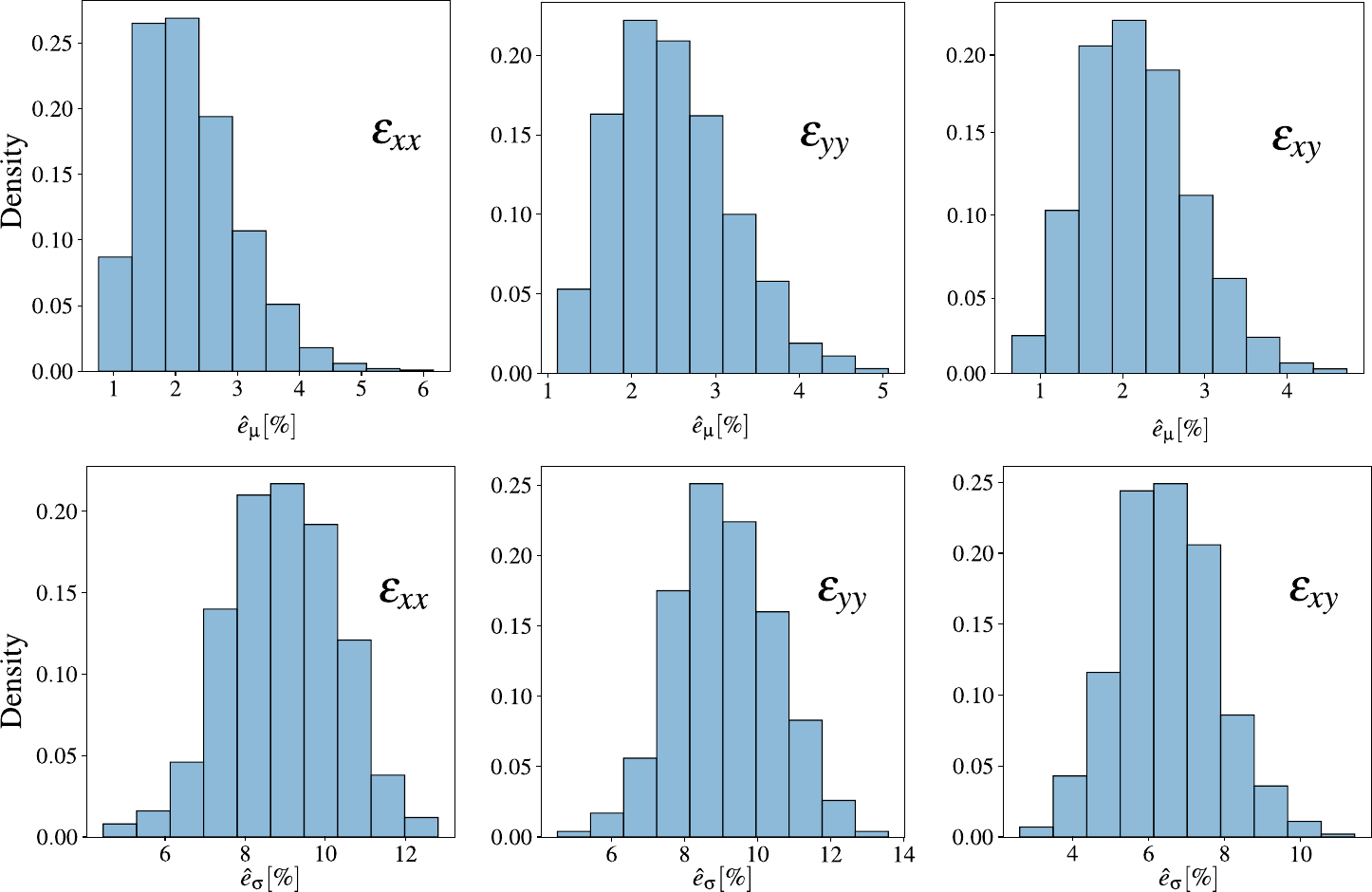}
	\caption{Histogram of normalized error metrics for the mean (top row) and standard deviation (bottom row) for the clamped plate case where $t \sim \mathcal{U}[7, 10] \ \text{mm}$}
	\label{fig: Fig 5}
	
\end{figure}

\begin{figure}[htp!]

	\centering
	\includegraphics{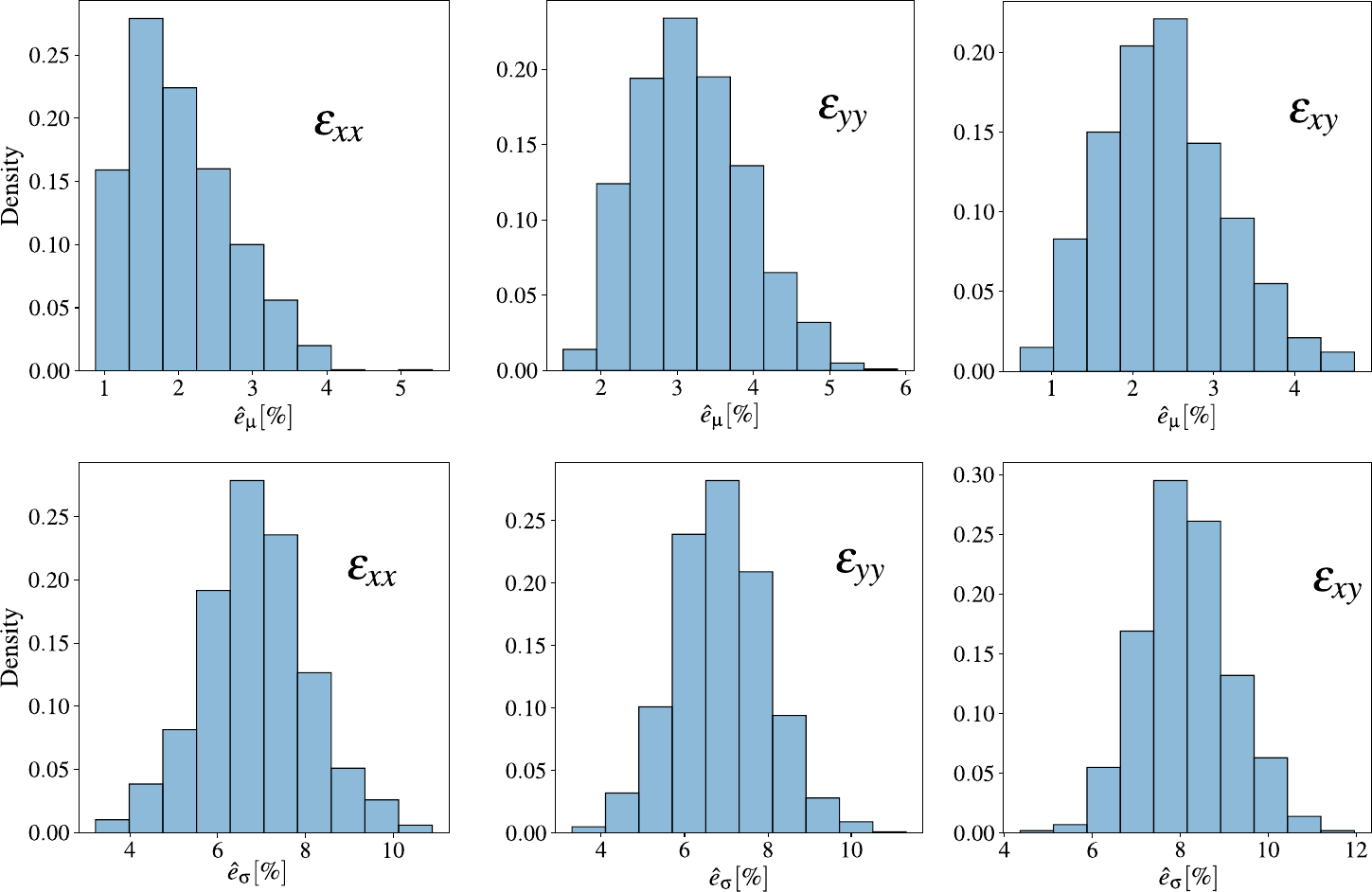}
	\caption{Histogram of normalized error metrics for the mean (top row) and standard deviation (bottom row) for the clamped plate case where $t \sim \mathcal{U}[8, 10] \ \text{mm}$}
	\label{fig: Fig 6}
	
\end{figure}

\begin{figure}[htp!]

	\centering
	\includegraphics[scale=0.9]{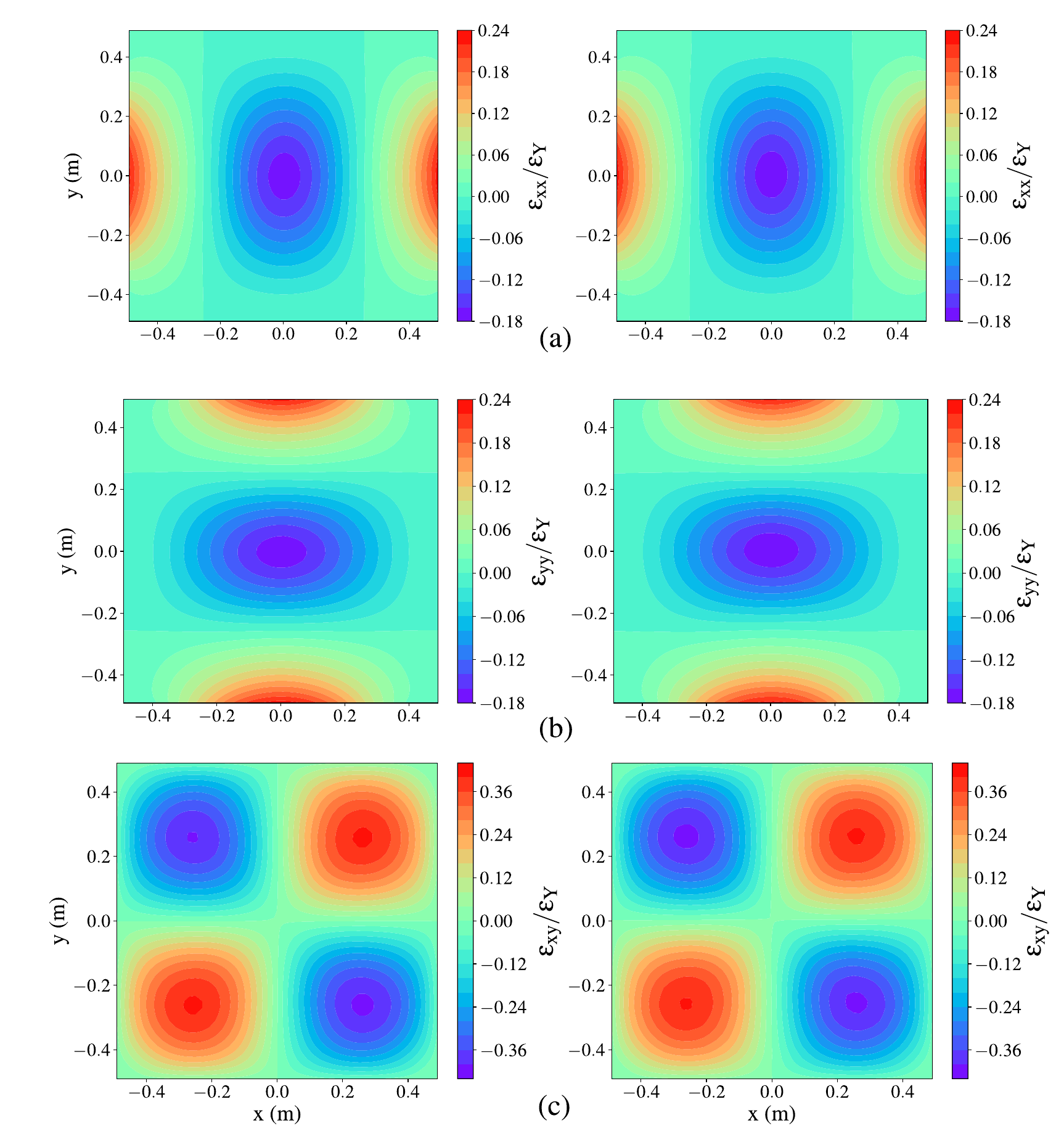}
	\caption{Contour plot of non-dimensionalized strain response mean field on test set instances for (a) $\varepsilon_{\text{xx}}$, (b) $\varepsilon_{\text{yy}}$, (c) $\varepsilon_{\text{xy}}$ predicted by the FE model (left column) and the surrogate (right column) for the clamped plate case}
	\label{fig: Fig 7}
	
\end{figure}

\begin{figure}[htp!]

	\centering
	\includegraphics[scale=0.9]{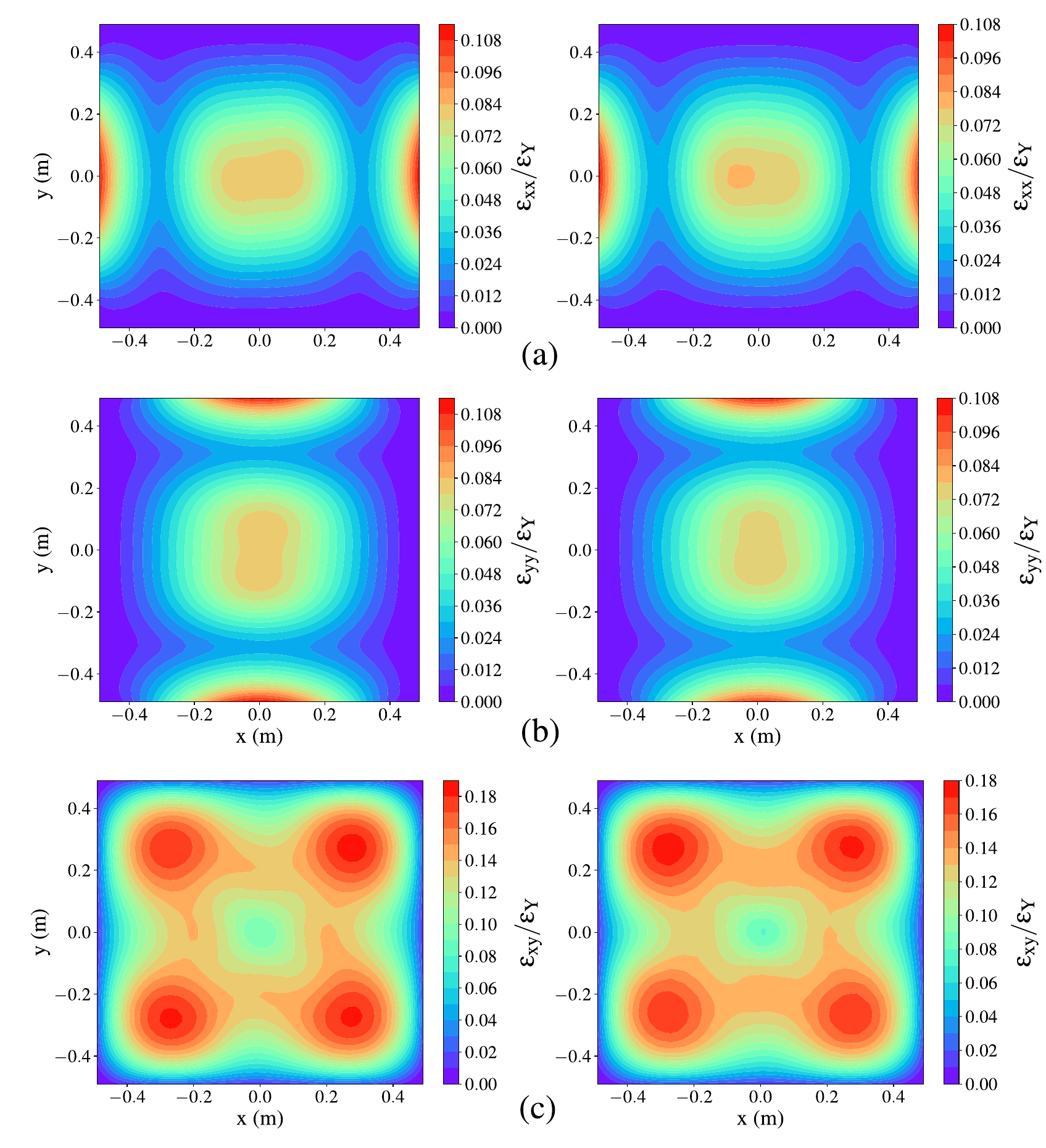}
	\caption{Contour plot of non-dimensionalized strain response standard deviation field on test set instances for (a) $\varepsilon_{\text{xx}}$, (b) $\varepsilon_{\text{yy}}$, (c) $\varepsilon_{\text{xy}}$ predicted by the FE model (left column) and the surrogate (right column) for the clamped plate case}
	\label{fig: Fig 8}
	
\end{figure}

Comparing the results between the two cases indicates that they exhibit equivalent accuracy in terms of the mean strain response but in terms of the standard deviation the case where $t \in [7, 10] \ \text{mm}$ was found to have decreased performance. The mean strain response fields evaluated for the test set according to Eq. \ref{eq17} and \ref{eq19} for that case are plotted in Figure \ref{fig: Fig 7} for all three strain components.  Each row corresponds to a specific component while the left column contains the results from the exact (FE) model and right one for the surrogate. The same results, this time for the standard deviation calculated according to Eq. \ref{eq18} and Eq. \ref{eq20} are plotted in Figure \ref{fig: Fig 8}. All results have been normalized with respect to their corresponding yield strain, i.e., normal and shear, and are presented in dimensionless form.

\clearpage

\begin{figure}[htp!]

	\centering
	\includegraphics[scale=0.88]{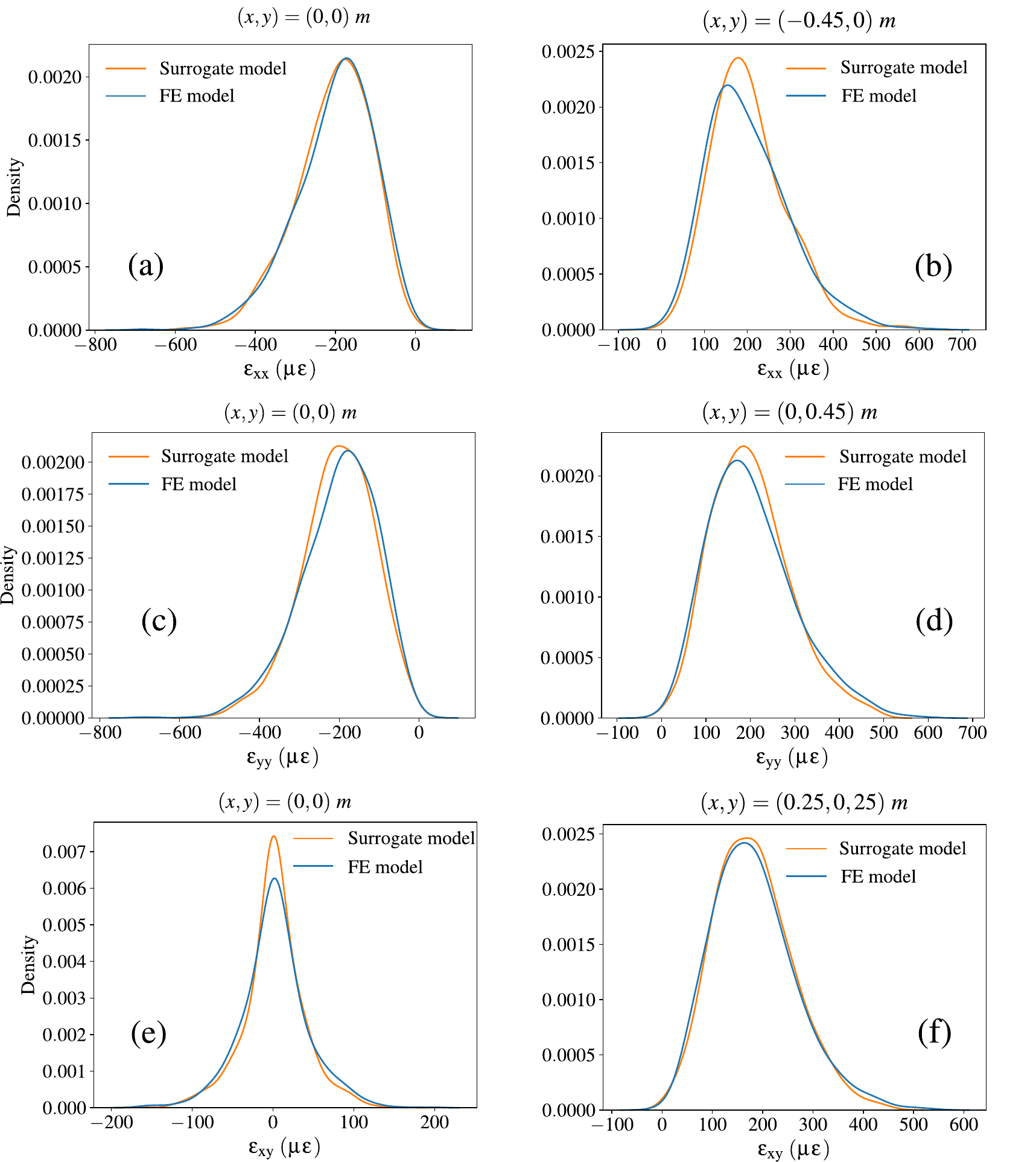}
	\caption{Marginal pdfs of strain response quantities for $\varepsilon_{\text{xx}}$ (a) \& (b), $\varepsilon_{\text{yy}}$ (c) \& (d), $\varepsilon_{\text{xy}}$ (e) \& (f) at different locations over the clamped plate expressed using the local coordinate system.}
	\label{fig: Fig 9}
	
\end{figure}

These plots confirm that the surrogate achieves a very accurate approximation in terms of the mean field behavior but exhibits decreased accuracy in the standard deviation. The error levels of the former are acceptable within the context of this analysis, while the latter was observed to be affected by the existence of significant outliers. This is evident in the marginal strain response probability density functions (PDFs) showcased in Figure \ref{fig: Fig 9}, which feature significantly heavy tails. These PDFs correspond to different strain components and locations over the plate and were fitted from the instances of the test set using kernel density estimation (KDE). This behavior can be attributed to the choice of a unit spherical prior and a diagonal Gaussian approximate posterior which are known to favor mean field approximations and are known to sometimes fail to satisfactorily encode more complex distributional patterns \cite{Yang2019}.

The cause of these outliers is the load profile, which is itself skewed and propagates this distributional shape to the response, due to the linearity of the input-output system. The presence of heavy tails is also the cause behind the discrepancy in the standard deviation errors between the two considered cases. This can be attributed again to the linearity of the input-output system, which dictates that lower thickness levels would lead to higher strain response, thus accentuating the effect of the outliers. It should be noted that several mitigating strategies have been proposed in the literature that could potentially enable this type of model to better represent more complex distributions (e.g., \cite{Rezende2015, Burda2015}). The authors considered investigating this out of the scope of the present work and have thus reserved it for future research. Additionally, more sophisticated techniques for hyperparameter tuning, which were not investigated within the context of this work, could potentially lead to improved performance. Finally, in terms of computational cost, to obtain an equal number of 3000 samples the trained surrogate requires a total of 3.2 seconds, which counting the offline cost amounts to an 88\% decrease in the overall required computational time. It should be mentioned here that surrogate calculations utilize GPU acceleration, while the same is not the case for the FE model.

\clearpage

\subsection{Ship hull structural element under stochastic loading}
\label{sec42}

\begin{figure}[htp!]

	\centering
	\includegraphics[scale=1.0]{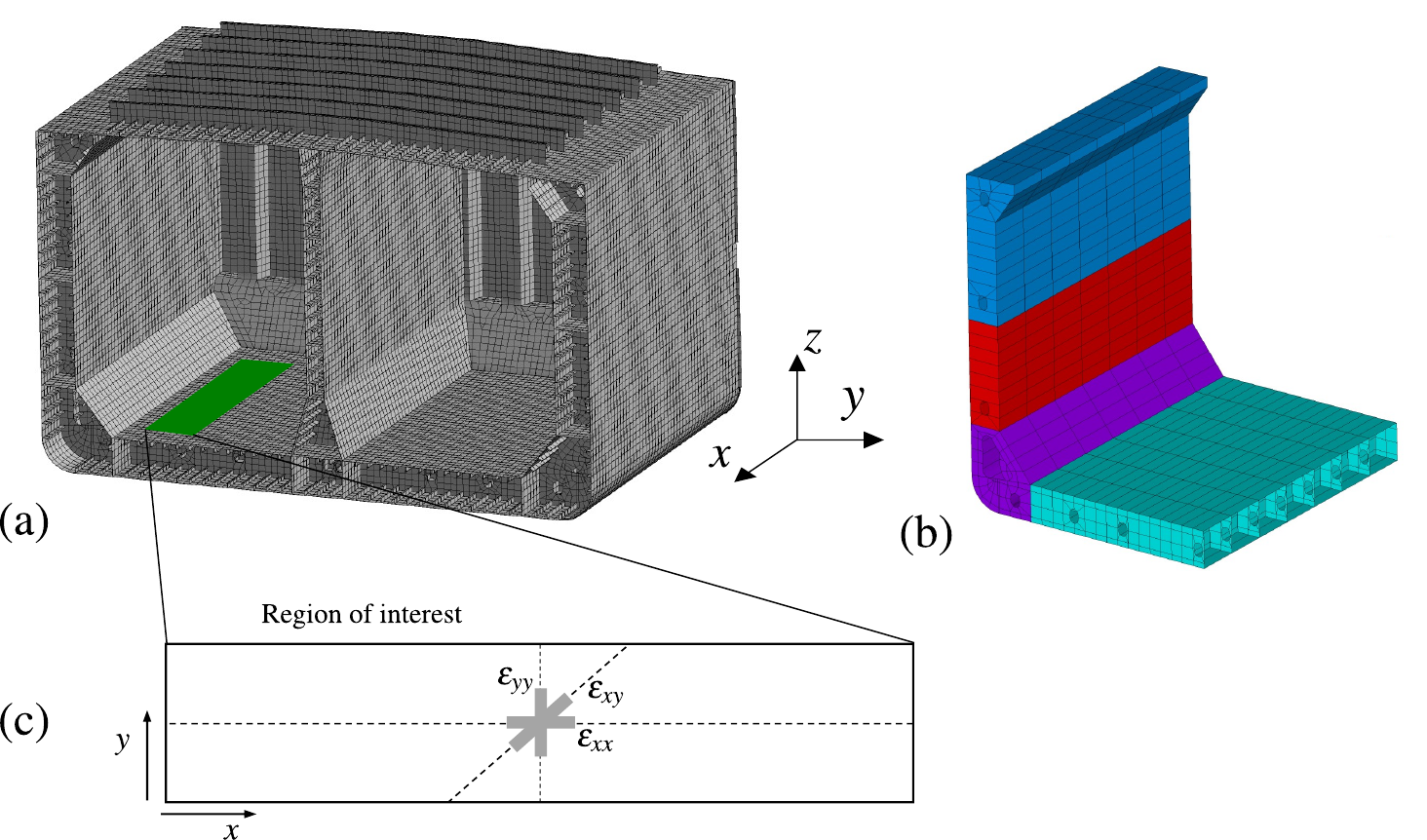}
	\caption{Interior view of three-hold compartment FE model (a), highlighted regions with potential thickness loss (b) and schematic representation of region of interest for the surrogate model (c)}
	\label{fig: Fig 10}
	
\end{figure}

A more complex structure was considered as the second case study, namely a ship hull. Ship hulls are characterized by a complex structural arrangement consisting of a large number of structural elements. They are also exposed to significant operational variability due to the stochastic nature of the ocean environment, uncertainties related to the structure itself (e.g., material properties), as well as the influence of the human factor. Structural simulations that employ high-fidelity models of the entire structure, or large part thereof, are associated with significant computational cost, which can become prohibitive when a large number of simulations is required. However, accounting for the influence of basic random variables that describe the different sources of uncertainty dictates that a probabilistic treatment is required. Furthermore, the complexity of the structure itself makes the use of global models desirable, even when only localized responses are needed, as they can more efficiently account for the complex interactions between different structural elements without the need for the simplifying assumptions that would be necessary for local models. 
The proposed surrogate can address these two aspects simultaneously, while drastically reducing computational requirements. To illustrate this, a high-fidelity three-compartment FE model of a product carrier has been generated, a part of which is shown indicatively in Figure \ref{fig: Fig 10} (a), demonstrating the interior of one of the three modeled holds. The model was discretized using a combination of linear shell and beam elements, leading to approximately $890\cdot10^3$ unconstrained DOFs, which correspond to an equal dimension of the square matrix to be inverted to obtain a solution. This indicates the significant wall-time required to obtain that solution, even for a linear elastic analysis such as the one considered herein, which takes approximately 1 minute.

In the considered problem, the structure is subjected to hydrostatic pressure on its external surface while internally it is subjected to pressure due to the liquid cargo it carries. It has been found that the latter is stochastic in nature due to the human influence in controlling the quantity of cargo in each hold. Namely, it has been observed that the cargo hold filling rates follow a beta distribution \cite{Garre2009}, which was assumed for this analysis as well. Its parameters were calculated so as to correspond to a population mean of 95\% and a standard deviation of 1.4\%. The filling rates over the six holds included in the model were treated as independent and identically distributed (i.i.d.) random variables. The resultant pressure distribution is tightly centered around a nominal value, thus exhibiting significantly lower variability than the clamped plate case. Considering that the implementation of the surrogate is the focus of this work, for more details on the modeling process the reader is referred to Silionis et al. \cite{Silionis2023}.

For the purposes of this work, a specific region of interest was isolated from the global model, indicated with green in Figure \ref{fig: Fig 10} (a), which is located on the inner bottom plating. The placement of the region is again related to the intended use of the surrogate in SHM applications and specifically corrosion-induced thickness loss monitoring. This region is significantly affected by corrosion, as it is adjacent to tanks that carry seawater. In Figure \ref{fig: Fig 10} (b) each color highlights a region around a particular tank that has been assumed to exhibit uniform thickness loss. This schematic corresponds to one symmetric half of the model, but the assumption is made that the same levels of thickness loss occur in the other half as well. The thickness loss levels over these regions, which do not have the same initial thickness, constitute the vector of conditioning variables of the problem, namely $t_l \in \mathbb{R}^4$.

As in the previous case, in order to obtain training data an MCS is required in order to propagate the load-related uncertainty and at the same time generate solutions for different realizations of the conditioning variable vector. Due to the more significant computational cost of the three-hold compartment model, this time $N_{\text{MC}} = 1000$ realizations were obtained from the six i.i.d. filling rates and the conditioning variables using LHS. The latter were also drawn independently from the same uniform distribution, namely $t_l^{(i)} \sim \mathcal{U}[0,2] \ \text{mm}$, where $t_l^{(i)}$ refers to a component of the thickness loss vector. The range of the uniform distribution was selected based on available thickness loss measurements from existing vessels of the same type, for a specific time in their operational life \cite{Paik2003}. It should be noted that although the same distribution is selected for every thickness loss term, they are still sampled independently. This conforms to the logical assumption that the different regions do not exhibit the same thickness loss at the same time.

Overall, extracting the strain response field required 16 hours using the same machine as in the other case. This time, the generated strain response fields were $8 \times 24$, with the lower dimensionality being a result of the scale of the FE model which dictated a coarser discretization. The same strain components, namely $\varepsilon_{\text{xx}}$, $\varepsilon_{\text{yy}}$, $\varepsilon_{\text{xy}}$, were extracted and are shown indicatively in Figure \ref{fig: Fig 10} (c), along with the local coordinate system of the region of interest, which is consistent with the global one shown in the same schematic. This time, a 70\%-30\% split was employed to separate the data into training and test sets and min-max scaling based on the training set was used to normalize them. A similar heuristics-based process was followed for hyperparameter selection, ultimately leading to a shallower architecture, as shown in Table \ref{tab:Tab2}.

Compared to the clamped plate case where the model has approximately $3.5 \times 10^6$ trainable parameters, in this case the number drops to approximately $2 \times 10^5$. trainable parameters, in this case the number drops to approximately $2\times10^5$. Furthermore, the dimensionality reduction occurs in a more rapid fashion in this architecture, while a 2-dimensional latent space was found to be sufficient, with higher-dimensional alternatives yielding no tangible improvement in performance. The same can be said about deepening the architecture or utilizing a larger number of kernels to produce deeper feature maps. Ultimately, this architecture benefitted from training for a greater number of epochs than the clamped plate case, namely 1000 using again the Adam optimizer, a learning rate this time equal to $5 \times 10^{-4}$ and a batch size of 8. Due to the vastly lower number of trainable parameters training required 20 minutes for each model, again using GPU acceleration.

\begin{table}[htp!] 
  \centering
  \caption{Layout of CNN-CVAE architecture for the ship hull structural element problem}
    \begin{tabular}{cp{8.8em}p{16.35em}p{6.75em}}
    \toprule
          & Layer type & Layer Parameters & Activation \\
    \midrule
    \multicolumn{1}{c}{\multirow{6}[2]{*}{Encoder}} & 2D Convolutional & 32 (3×5) kernels with (1×1) stride & Leaky ReLU \\
          & 2D Convolutional & 32 (3×5) kernels with (2×2) stride & Leaky ReLU \\
          & 2D Convolutional & 32 (3×5) kernels with (1×1) stride & Leaky ReLU \\
          & Flattening layer & -     & - \\
          & Fully Connected & Weights: (1024 × 64) & Leaky ReLU \\
          & Fully Connected & Weights: (64 × 2) & Leaky ReLU \\
    \midrule
    \multicolumn{1}{c}{\multirow{6}[2]{*}{Decoder}} & Fully Connected & Weights: (6 × 64) & Leaky ReLU \\
          & Fully Connected & Weights: (64 × 1024) & Leaky ReLU \\
          & Reshaping layer & -     & - \\
          & 2D Deconvolutional & 32 (3×5) kernels with (1×1) stride & Leaky ReLU \\
          & 2D Deconvolutional & 32 (3×5) kernels with (2×2) stride & Leaky ReLU \\
          & 2D Deconvolutional & 1 (3×5) kernels with (1×1) stride & Sigmoid \\
    \bottomrule
    \end{tabular}
    \label{tab:Tab2}
\end{table}%

\begin{figure}[htp!]

	\centering
	\includegraphics[scale=1.0]{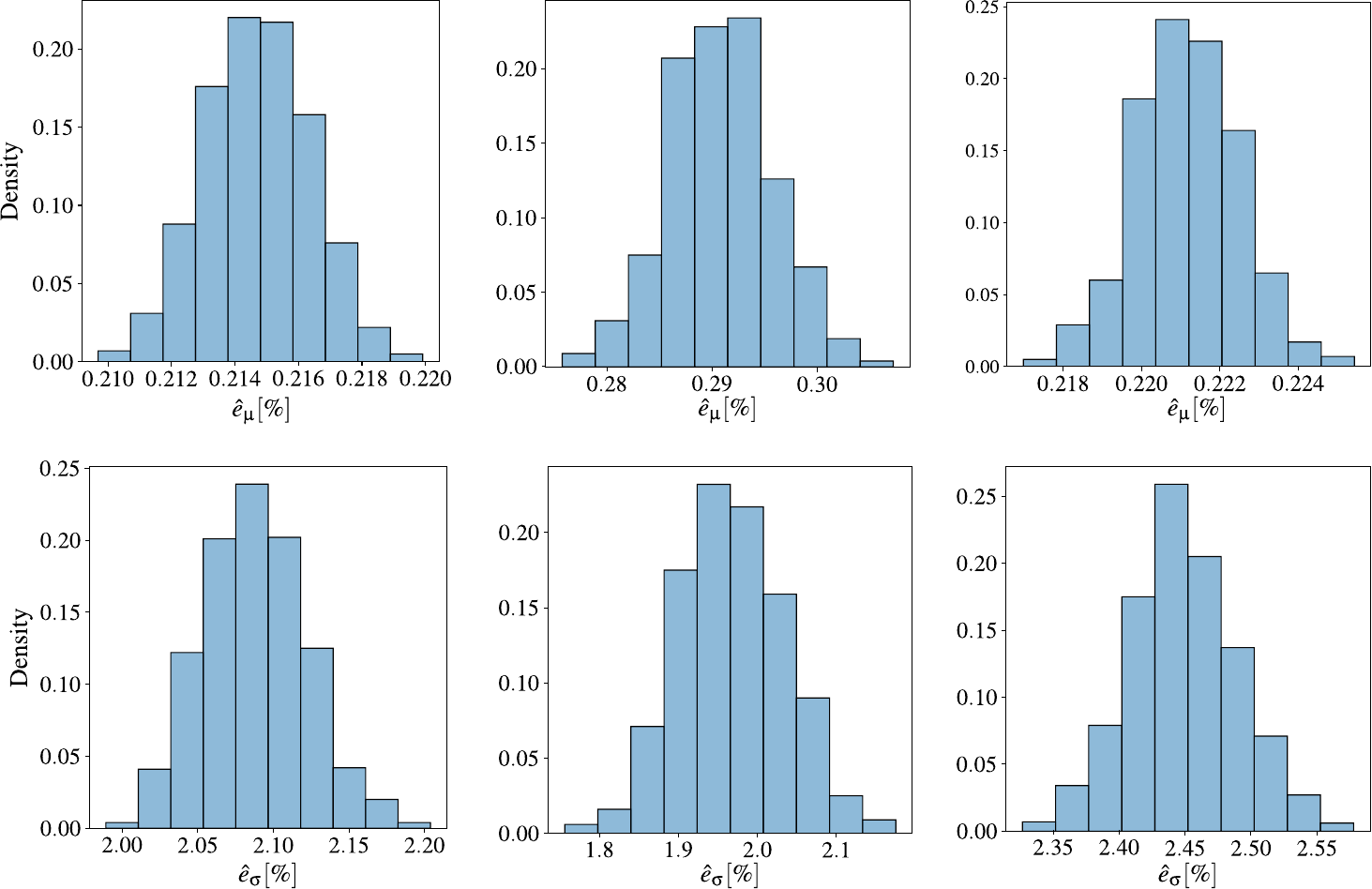}
	\caption{Histogram of normalized error metrics for the mean (top row) and standard deviation (bottom row) for the hull structural element case}
	\label{fig: Fig 11}
	
\end{figure}

\begin{figure}[htp!]

	\centering
	\includegraphics[scale=0.9]{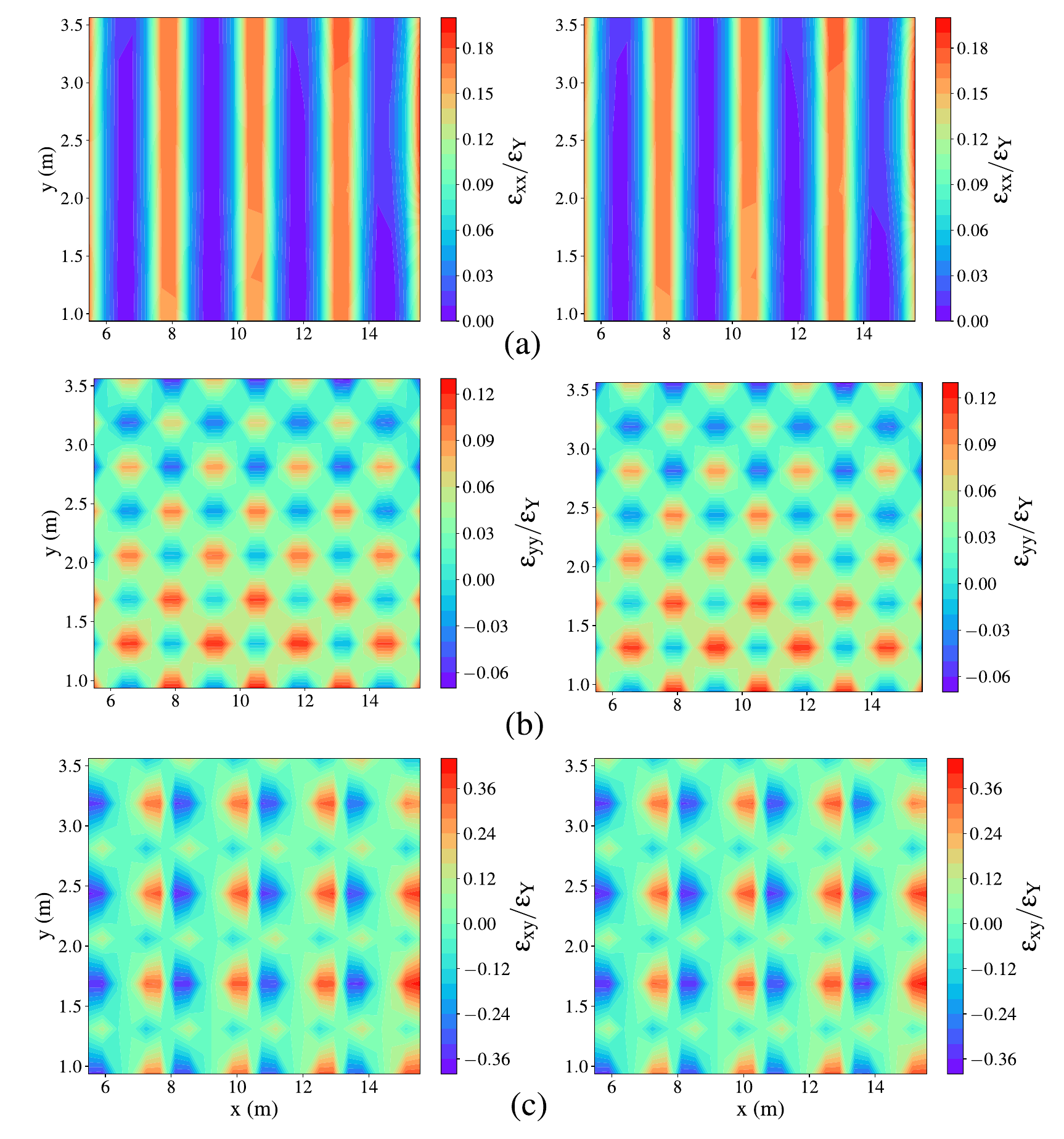}
	\caption{Contour plot of non-dimensionalized strain response mean field on test set instances for (a) $\varepsilon_{\text{xx}}$, (b) $\varepsilon_{\text{yy}}$, (c) $\varepsilon_{\text{xy}}$ predicted by the FE model (left column) and the surrogate (right column) for the hull structural element case}
	\label{fig: Fig 12}
	
\end{figure}

\begin{figure}[htp!]

	\centering
	\includegraphics[scale=0.9]{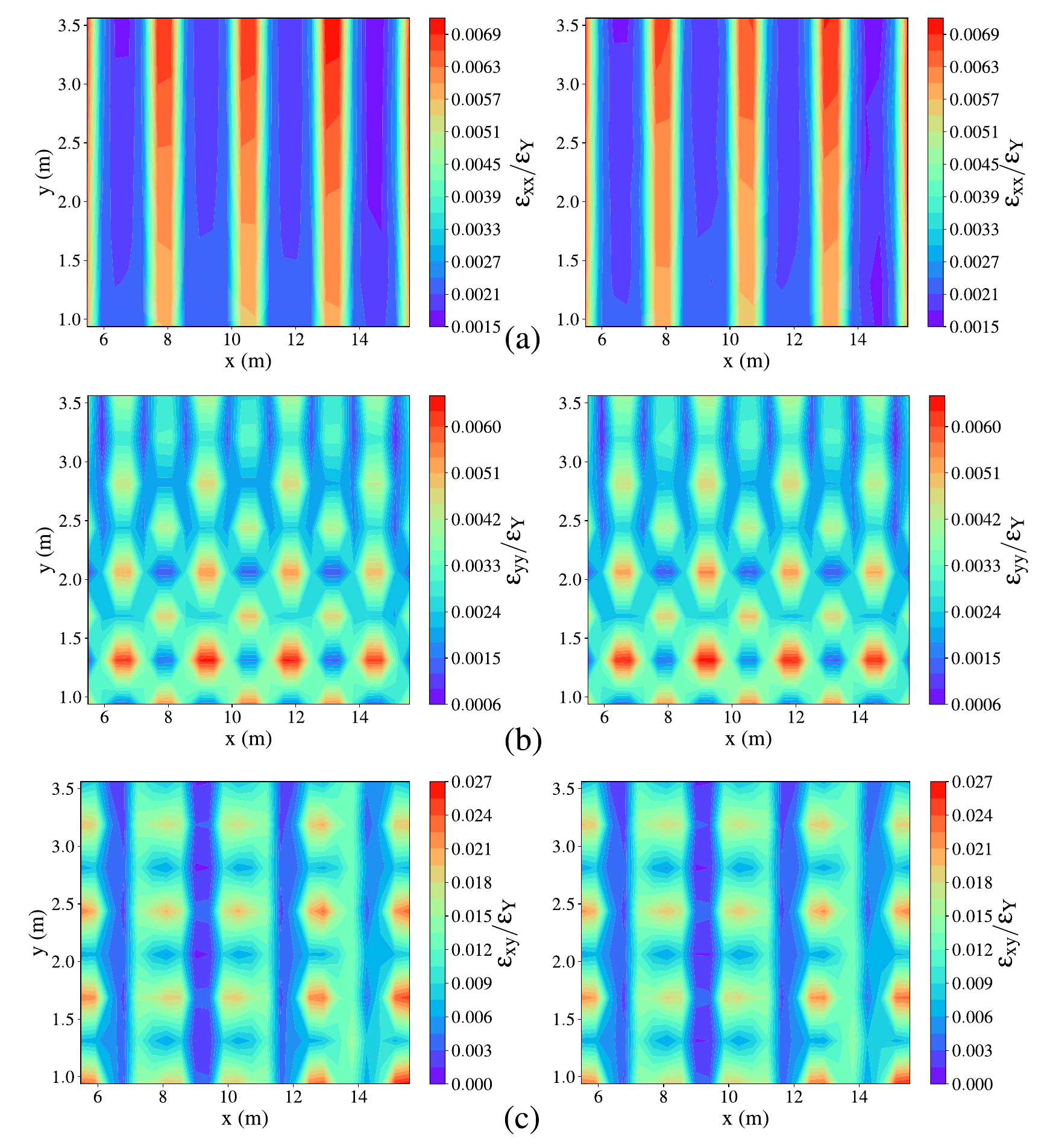}
	\caption{Contour plot of non-dimensionalized strain response standard deviation field on test set instances for (a) $\varepsilon_{\text{xx}}$, (b) $\varepsilon_{\text{yy}}$, (c) $\varepsilon_{\text{xy}}$ predicted by the FE model (left column) and the surrogate (right column) for the hull structural element case}
	\label{fig: Fig 13}
	
\end{figure}

The same process as in the clamped plate problem was employed to assess the CNN-CVAE performance by performing an MCS with $N_{\text{MC}} = 1000$ samples from the latent space distribution used to calculate the error metrics. The obtained histograms are provided in Figure \ref{fig: Fig 11} and indicate the high level of performance achieved by the model as the mean and standard deviation errors are well within acceptable limits. This is also evident by the contour plots of the strain response mean and standard deviation fields, presented in Figure \ref{fig: Fig 12} and \ref{fig: Fig 13}.

\begin{figure}[htp!]

	\centering
	\includegraphics[scale=0.9]{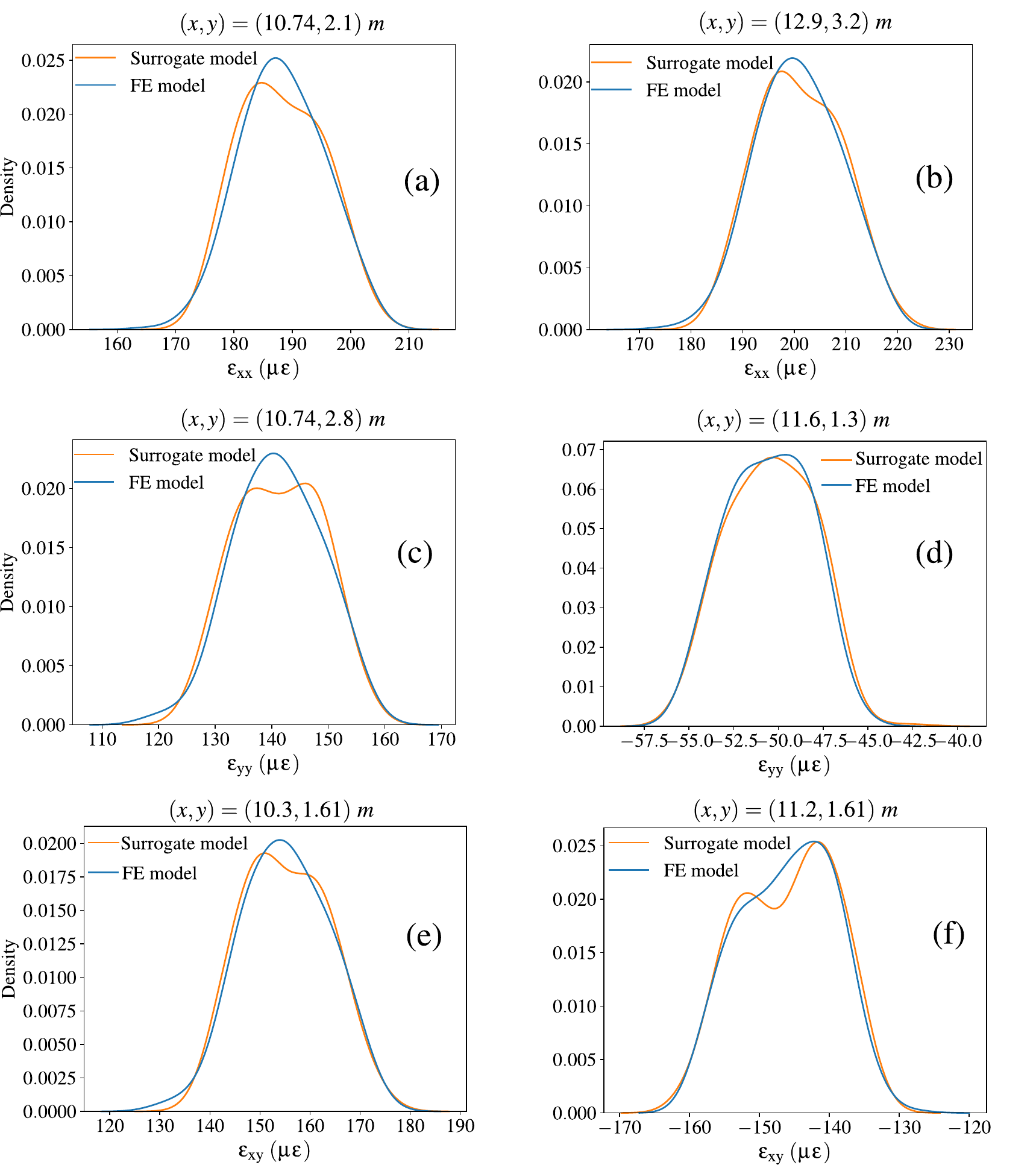}
	\caption{Marginal pdfs of strain response quantities for $\varepsilon_{\text{xx}}$ (a) \& (b), $\varepsilon_{\text{yy}}$ (c) \& (d) and $\varepsilon_{\text{xy}}$ (e) \& (f) at different locations over the hull region of interest expressed in the local coordinate system}
	\label{fig: Fig 14}
	
\end{figure}

An important observation in this case is that the variance of the error metrics is significantly lower than in the clamped plate case. This behavior can be attributed to the low levels of uncertainty contained in the structural response data, in turn attributed to the characteristics of the beta distribution governing the pressure loads. Due to this, the primary source of uncertainty is the conditioning vector, which being uniformly distributed, is well suited to the mean-field approximation capabilities offered by the employed model. This is illustrated even more clearly in Figure \ref{fig: Fig 14} which plots marginal strain response PDFs, obtained using KDE, for each strain component at specific locations over the region of interest. Indeed, it is evident that the distribution shapes are symmetric and do not exhibit significant variance, which conforms well with the representation capabilities of the employed model. It should be noted that the locations are expressed in the region’s local coordinate system indicated in Figure \ref{fig: Fig 10} (c).

The less complex representation that needs to be encoded in the latent space combined with the lower dimensionality of the problem are the contributing factors that explain both the higher accuracy in this case, as well as the reduced computational requirements, i.e., number of trainable parameters. The latter is also reflected in the fact that obtaining 1000 samples from the trained surrogate requires a mere 0.19 seconds using GPU acceleration. Accounting for the offline cost, i.e., training time, this amounts to a 97.9 \% decrease in the required computational time.
Finally, it is worth pointing out that in this case, the surrogate offers certain advantages compared to computational models of equivalent cost. Namely, by virtue of its training through the global model response and its inherent uncertainty quantification and representation capabilities, it can efficiently encode the effects that different thickness loss levels at neighboring regions have on the region interest. In a traditional setting where a local model would be employed, these would have to be modeled through the imposed boundary conditions. Considering the complexity of the structure and the loading this would be a highly challenging task, which the proposed surrogate is inherently capable of achieving.

\clearpage

\clearpage

\clearpage

\clearpage

\section{Concluding remarks}
\label{sec5}

In this work, a probabilistic surrogate modeling architecture for high-dimensional stochastic structural simulations was proposed. It was based on conditional deep generative models, namely a CNN-CVAE, and its aim was to provide computationally inexpensive alternatives to FE-based MC simulations, which at the same time offer high reconstruction accuracy and uncertainty quantification. The surrogate model was applied on two cases motivated by its potential application within SHM schemes, where the goal was to reconstruct spatial field strain responses under stochastic loading, conditioned on specific structural geometry parameters.

Results indicated that the proposed surrogate can effectively quantify the load-related uncertainty through efficiently encoding it within latent model variables, although it can be ineffective in cases where it is characterized by significant complexity. This was evident in the clamped plate case, where it exhibited reduced accuracy when the structural response data contained significant outliers. The root cause of this was considered to be the choice of a unit spherical Gaussian prior over the latent variables and a corresponding diagonal Gaussian approximate posterior, which are more effective in mean field approximations, as indicated in the hull structural element case. Although not explored within this work, a different choice that allows for more expressive latent representations could mitigate the problem. In terms of computational cost, the surrogate achieves a significant reduction compared to FE-based simulations. Considering that heuristics were employed to determine the model hyperparameters, it is conceivable that a further reduction can be achieved if a hyperparameter optimization process that is aimed at this task is employed. The same could also be geared towards further improving performance, or towards a goal that combines both.

Although applied on a hull structural element, the proposed surrogate can be applied to any large-scale structure that operates under uncertainty and where a localized response is the desired output from a global model which is associated with high computational cost. One typical such application is sub-modeling or global-local modeling, where the response quantity of interest would be nodal displacements. Furthermore, the surrogate is flexible in the architecture of the encoder and decoder, in the sense that although convolutional layers are an appealing choice for spatial field problems, if that is not the case then fully connected layers can be used without any change to the method’s working principle. Although particularly suited to problems where latent uncertainties are present, the proposed surrogate architecture can also be employed in more typical applications where its ordered latent structure can lead to improved performance. Finally, although applied for static structural simulations, the proposed architecture could conceivably be employed for time-dependent applications to learn latent representations of solution field snapshots, that could then be fed to a model capable of handling sequential data to capture the time dependence.

\clearpage

\bibliographystyle{IEEEtran}
\bibliography{references}

\end{document}